\documentclass[sigconf]{acmart}

\AtBeginDocument{%
  \providecommand\BibTeX{{%
    \normalfont B\kern-0.5em{\scshape i\kern-0.25em b}\kern-0.8em\TeX}}}

\copyrightyear{2024}
\acmYear{2024}
\setcopyright{acmlicensed}\acmConference[KDD '24]{Proceedings of the 30th ACM SIGKDD Conference on Knowledge Discovery and Data Mining}{August 25--29, 2024}{Barcelona, Spain}
\acmBooktitle{Proceedings of the 30th ACM SIGKDD Conference on Knowledge Discovery and Data Mining (KDD '24), August 25--29, 2024, Barcelona, Spain}
\acmDOI{10.1145/3637528.3672059}
\acmISBN{979-8-4007-0490-1/24/08}

\usepackage{multirow}

\usepackage{algorithm}
\usepackage{algorithmic}

\usepackage{hyperref}
\usepackage{subfigure}

\begin{document}

\title{Offline Imitation Learning with Model-based Reverse Augmentation}

\author{Jie-Jing Shao}
\authornote{Both authors contributed equally to this research.}
\affiliation{%
  \institution{National Key Laboratory for Novel Software Technology}
  \city{Nanjing University, Nanjing}
  \country{China}
  \postcode{210023}
}
\email{shaojj@lamda.nju.edu.cn}

\author{Hao-Sen Shi}
\authornotemark[1]
\affiliation{%
  \institution{National Key Laboratory for Novel Software Technology}
  \city{Nanjing University, Nanjing}
  \country{China}
  \postcode{210023}
}
\email{shihs@lamda.nju.edu.cn}

\author{Lan-Zhe Guo}
\authornote{Corresponding Authors.}
\affiliation{%
  \institution{National Key Laboratory for Novel Software Technology\\
  School of Intelligence Science and Technology}
  \city{Nanjing University, Nanjing}
  \country{China}
  \postcode{210023}
}
\email{guolz@nju.edu.cn}

\author{Yu-Feng Li}
\authornotemark[2]
\affiliation{%
  \institution{National Key Laboratory for Novel Software Technology\\
  School of Artificial Intelligence}
  \city{Nanjing University, Nanjing}
  \country{China}
  \postcode{210023}
}
\email{liyf@nju.edu.cn}
\renewcommand{\shortauthors}{Jie-Jing Shao, Hao-Sen Shi, Lan-Zhe Guo, and Yu-Feng Li.}

\begin{abstract}
  In offline Imitation Learning (IL), one of the main challenges is the \textit{covariate shift} between the expert observations and the actual distribution encountered by the agent, because it is difficult to determine what action an agent should take when outside the state distribution of the expert demonstrations. 
  Recently, the model-free solutions introduce the supplementary data and identify the latent expert-similar samples to augment the reliable samples during learning. 
  Model-based solutions build forward dynamic models with conservatism quantification and then generate additional trajectories in the neighborhood of expert demonstrations. 
  However, without reward supervision, these methods are often over-conservative in the out-of-expert-support regions, because only in states close to expert-observed states can there be a preferred action enabling policy optimization. 
  To encourage more exploration on expert-unobserved states, we propose a novel model-based framework, called offline Imitation Learning with Self-paced Reverse Augmentation (SRA). 
  Specifically, we build a reverse dynamic model from the offline demonstrations, which can efficiently generate trajectories leading to the expert-observed states in a self-paced style.
  Then, we use the subsequent reinforcement learning method to learn from the augmented trajectories and transit from expert-unobserved states to expert-observed states. 
  This framework not only explores the expert-unobserved states but also guides maximizing long-term returns on these states, ultimately enabling generalization beyond the expert data. 
  Empirical results show that our proposal could effectively mitigate the covariate shift and achieve the state-of-the-art performance on the offline imitation learning benchmarks. 
  Project website: \url{https://www.lamda.nju.edu.cn/shaojj/KDD24_SRA/}. 
\end{abstract}

\ccsdesc[500]{Computing methodologies~Learning from demonstrations}
\ccsdesc[500]{Computing methodologies~Inverse reinforcement learning}

\keywords{Offline Imitation Learning, Model-based Imitation Learning, Offline Reinforcement Learning}



 \maketitle

\section{Introduction}

The recent success of offline Reinforcement Learning (RL) in various fields demonstrate the significant potential of addressing sequential decision-making problem in a data-driven manner~\cite{Tutorial_orl}. Offline reinforcement learning enables the learning of policies from the logged experience, reducing the reliance on online interactions and making reinforcement learning more practical, especially when online data collection may be expensive or risk-sensitive~\cite{DBLP:conf/corl/SinhaMG21, DBLP:conf/nips/QinZGCL0022, DBLP:conf/itsc/FangZGZ22}. 
However, in many real-world applications, designing the reward function, which determines the preferred agent behavior, is a critical prerequisite for offline reinforcement learning and can be prohibitively tricky. 
It usually needs to be custom-designed for each task. 
This requires sufficient prior knowledge and can be challenging in fields such as robotics~\cite{DBLP:journals/ijrr/BobuWTD22}, autonomous driving~\cite{DBLP:journals/ai/KnoxABSS23}, and healthcare~\cite{DBLP:journals/csur/YuLNY23}. 
One promising way to overcome this practical barrier is \textit{Offline Imitation Learning}, which trains policies directly from expert demonstrations without needing reward supervision. 

Offline imitation learning 
allows machine to implement sequential decision-making by imitating expert behaviors. It has achieved remarkable success on a variety of domains, such as robotic manipulation~\cite{DBLP:journals/ijira/FangJGXWS19, MimicPlay} autonomous driving~\cite{DBLP:journals/tits/MeroYDM22, DBLP:journals/tits/BhattacharyyaWPKMSK23} and language models~\cite{whitehurst1975language, DBLP:conf/nips/BrownMRSKDNSSAA20}. However, the effectiveness of imitation learning methods typically relies on the sufficient expert demonstrations. When the expert data is limited, the \textit{covariate shift} between expert observations and the actual distribution encountered by the agent makes imitation learning methods ineffective, both in theory~\cite{DBLP:conf/nips/RajaramanYJR20, DBLP:journals/pami/XuLY22} and practice~\cite{MILO, DWBC}. 
One naive solution to this problem is to collect more demonstrations from the expert, but this is costly and impractical in domains where expert experience is expensive and scarce, such as the robotics~\cite{MimicPlay} or healthcare~\cite{DBLP:conf/nips/JarrettBS20}. 

Recently, various advanced offline Imitation Learning methods have been proposed. They introduce the supplementary data, albeit sub-optimal, as the assistance to the limited expert data~\cite{DemoDICE,DWBC,OTIL,MILO,CLARE,BCDP,ILID}. 
Model-free methods typically identify the latent high-quality samples from supplementary. For examples, DemoDICE trains a sample discriminator via a regularized state-action distribution matching objective~\cite{DemoDICE}. DWBC assume the 
data come from a mixed distribution of expert policy and sup-optimal policy, and thus utilize positive-unlabeled learning to identify the expert-similar samples from the expert data~\cite{DWBC}. OTIL uses optimal transport to find an alignment with the minimal wasserstein distance between the supplementary trajectories and expert demonstrations~\cite{OTIL}. BCDP presents the idea to lead the agent from expert-unobserved states to expert-observed states and highlights that dynamic programming can efficiently use supplementary of low quality~\cite{BCDP}. A concurrent work ILID presents a similar idea and proposes a data selection method that identifies positive behaviors based on their resultant states, enabling explicit utilization of dynamics information from diverse demonstrations~\cite{ILID}. To efficiently utilize the supplementary, model-based methods build dynamic models from offline data with conservative quantification and then generate imaginary trajectories in areas similar to experts, expanding the offline dataset. MILO extends the adversarial imitation learning to utilize the samples from model-based augmentation~\cite{MILO}. CLARE estimates rewards for the demonstrations via dynamics conservatism and expert similarity, encouraging 
RL to keep the agent in the corresponding area~\cite{CLARE}. 

\begin{figure}[t]
  \centering
  \subfigure[Expert Trajectories]{
    \centering
    \includegraphics[width=0.48\columnwidth]{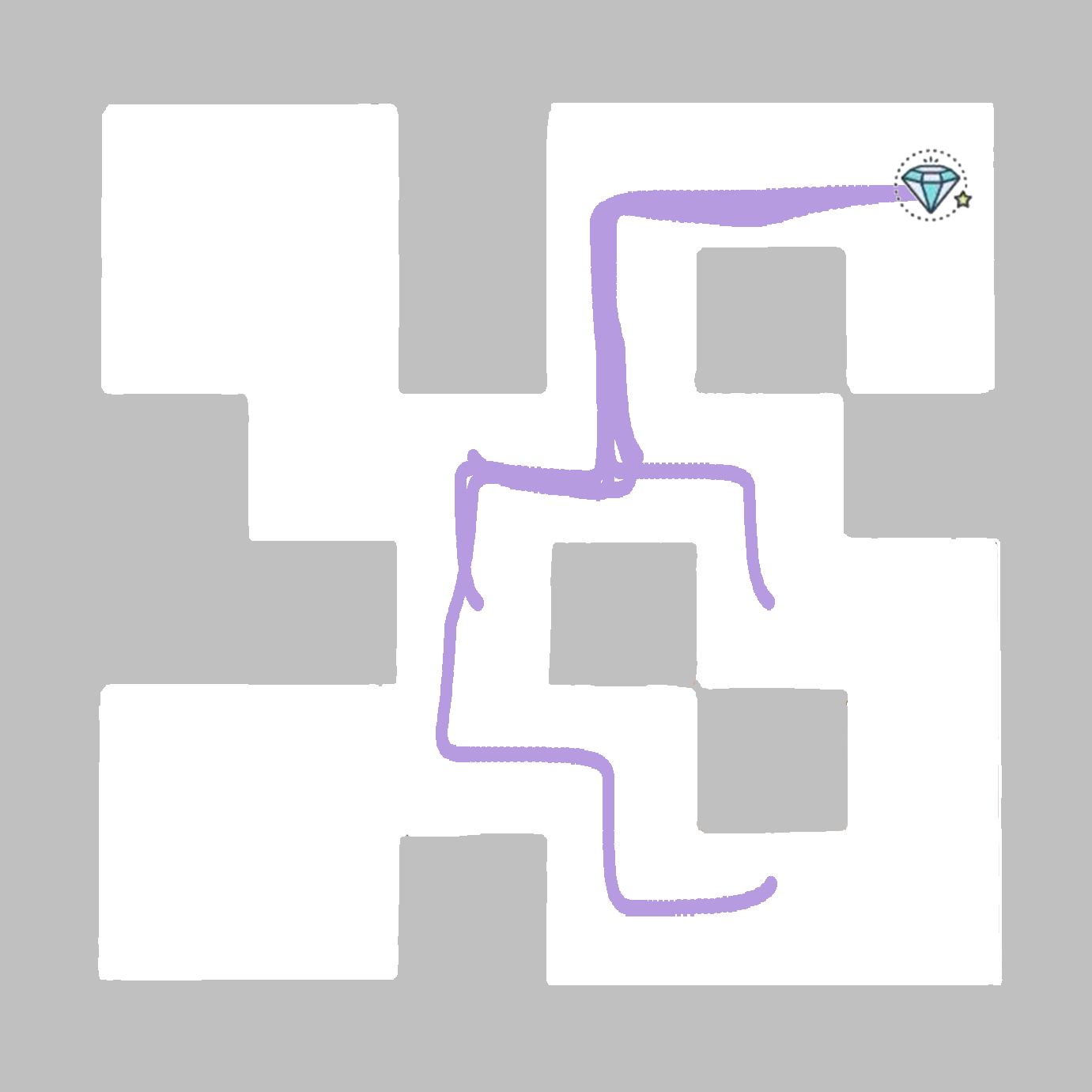}
  }
  \hspace{-0.1in}
  \subfigure[Cumulative return of MILO~\cite{MILO}]{
    \centering
    \includegraphics[width=0.48\columnwidth]{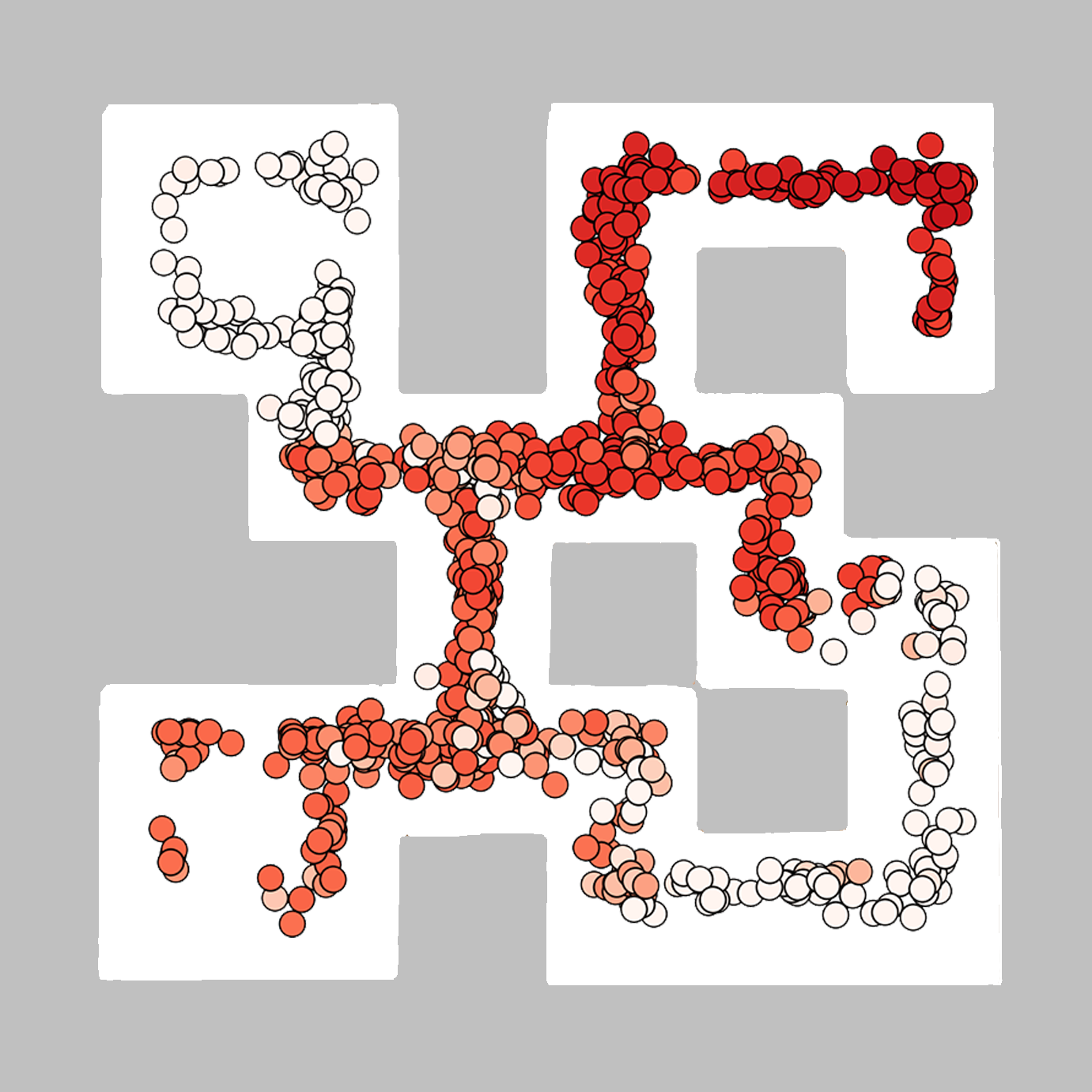}
  }
  \caption{Covariate shift problem. We visualize the expert trajectories and the state-wise cumulative return of MILO agent. Deeper red represents higher cumulative returns. 
  The agent performs well in the neighborhood of expert-observed states but performs poorly in the rest of the states.} 
  \label{covariate_shift}
\end{figure}
However, because samples in expert dataset are limited, the similarity measure often suffer from under-diversity, leading the ineffective exploration on the out-of-expert regions. As shown in the Figure~\ref{covariate_shift}, although MILO has generated more samples through model-based rollout, improving the generalization of the policy. However, there are still a large number of out-of-expert states where the policy performs poorly. 
To encourage more exploration on expert-unobserved states, we propose a novel model-based framework, called offline Imitation Learning with Self-paced Reverse Augmentation (SRA). 
  Specifically, we build a reverse dynamic model from the offline demonstrations, which can efficiently generate trajectories leading to the expert-observed states in a self-paced style.
  Then, we use the subsequent reinforcement learning method to learn from the augmented trajectories and transit from expert-unobserved states to expert-observed states. 
  This framework not only explores the expert-unobserved states but also maximizes the long-term returns on these states, ultimately enabling generalization beyond the expert data. 
  Empirical results show that our proposal could effectively mitigate the covariate shift and achieve the state-of-the-art performance on the offline imitation learning benchmarks.

\section{Related Work}

This work mainly focus on learning from demonstrations without reward supervision. There are two main branches to address this problem, that is, \textit{Imitation Learning} and \textit{Inverse Reinforcement Learning}. In this section, we review them sequentially. 

\paragraph{Imitation Learning} 
Behavioral Cloning is one of the most well-validated offline Imitation Learning, which optimizes the policy via supervised learning~\cite{DBLP:conf/nips/Pomerleau88}. 
According to the learning theory in~\cite{DBLP:conf/nips/RajaramanYJR20}, the effectiveness of behavioral cloning is fundamentally constrained by the scale of the expert dataset. This limitation arises as it's unclear how the agent should respond when confronted with states not represented in the expert demonstrations, which is also well-known as the \textit{covariate shift} problem in the traditional machine learning topics~\cite{sugiyama2007covariate,shao2024open,JMLR'24:Sword++,DBLP:conf/nips/ShaoGYL22}. 
To address this issue, recent studies~\cite{DBLP:conf/nips/LiuZYSZZL21, DBLP:conf/iclr/SasakiY21, DemoDICE, DWBC} introduce supplementary data from offline policies to assist the limited expert dataset, which is similar to semi-supervised learning~\cite{DBLP:conf/icml/GuoZJLZ20,DBLP:conf/nips/GuoZWSL22} that uses cheaper unlabeled data as supplementary to the limited labeled data. 
For examples, \citet{DemoDICE} takes the distribution matching with offline data to provide a policy regularization. 
\citet{DBLP:conf/iclr/SasakiY21} use prediction confidence from old policy during the learning to weight the samples, eliminating the noisy
behaviors in the supplementary data. 
\citet{DWBC} assume the supplementary data from a mixture of expert distribution and sub-optimal distribution, and then utilize positive-unlabeled learning to build a discriminator and identify the latent expert-similar samples. 
\citet{BCDP} presents the idea to lead the agent from expert-unobserved states to expert-observed states and highlight that dynamic programming can efficiently use supplementary data of low behavioral quality to support imitation learning. 
A concurrent work ILID~\cite{ILID} presents a similar idea and proposes a data selection method that identifies positive behaviors based on their resultant states, enabling explicit utilization of dynamics information from diverse demonstrations~\cite{ILID}. 
\citet{MILO} propose a model-based method MILO to alleviate covariate shift. They built a forward dynamic model and extended adversarial imitation learning to identify and utilize samples in model-based rollouts that are similar to expert samples.

\begin{figure*}[tb]
  \centering
  \subfigure[Challenge (i)]{
    \centering
    \includegraphics[width=0.24\textwidth]{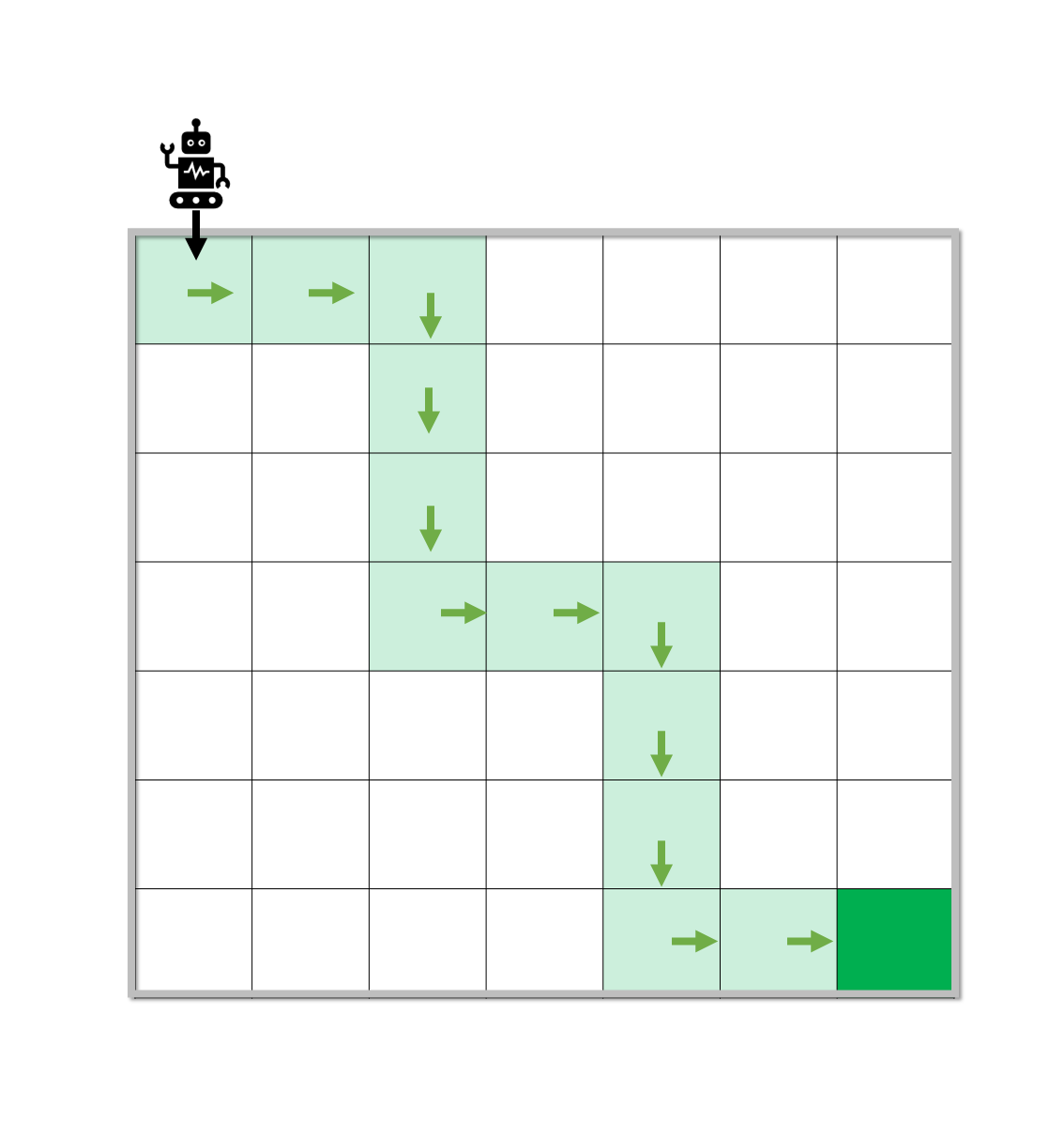}
    \label{m_1}
  }
  \hspace{-0.1in}
  \subfigure[Challenge (ii)]{
    \centering
    \includegraphics[width=0.24\textwidth]{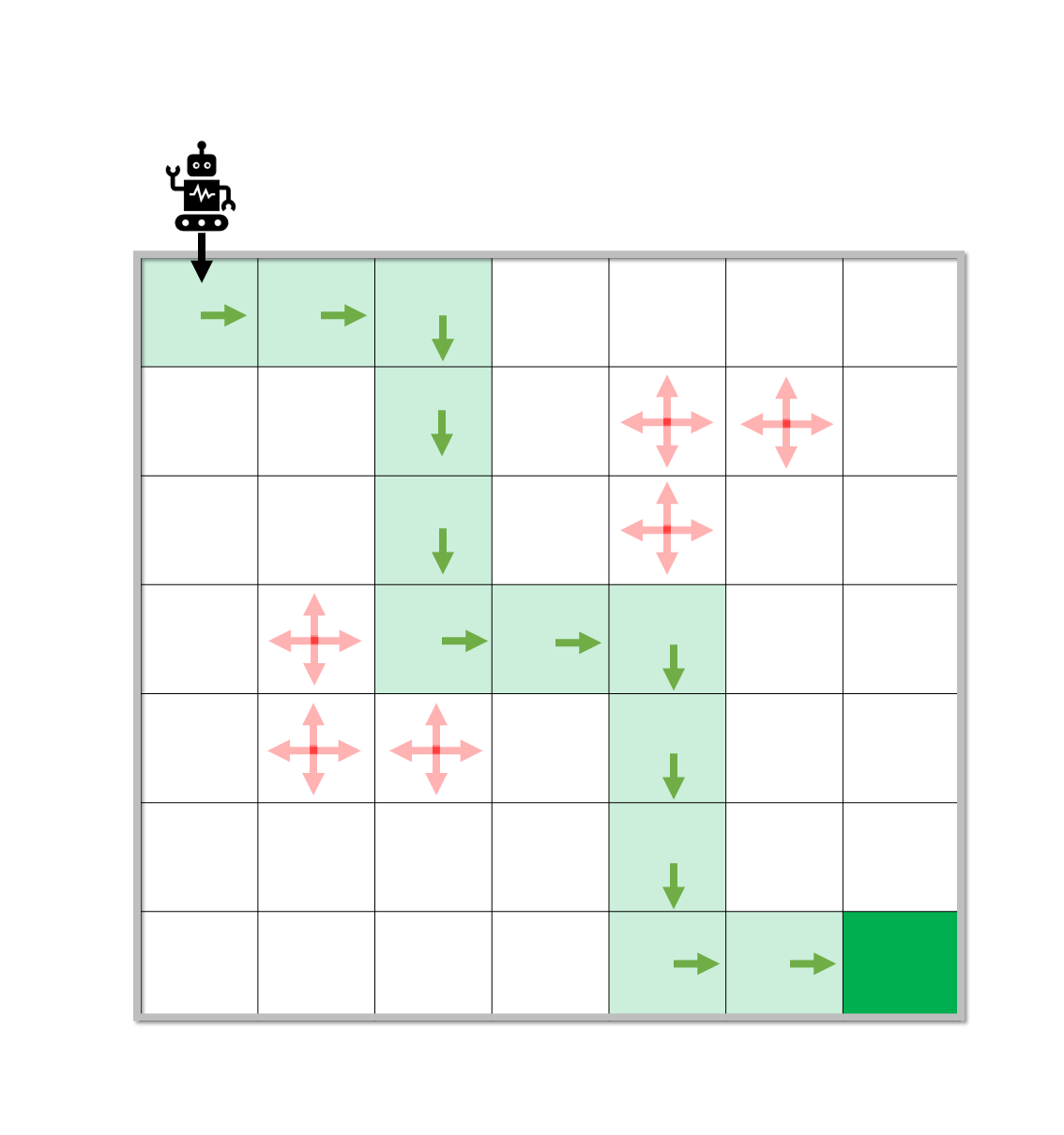}
  }
  \hspace{-0.1in}
  \subfigure[Challenge (iii)]{
    \centering
    \includegraphics[width=0.24\textwidth]{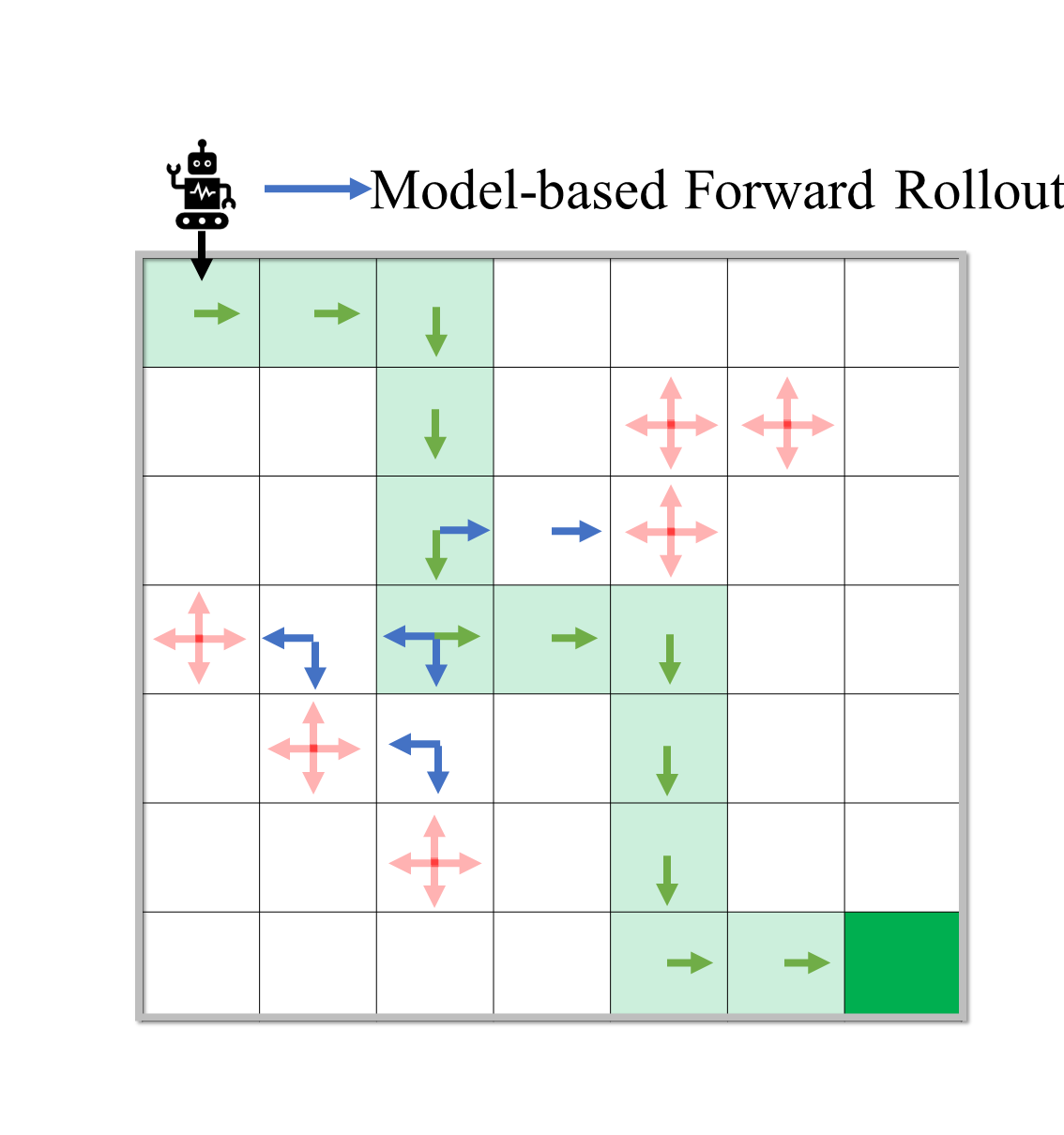}
  }
  \hspace{-0.1in}
  \subfigure[Our Proposal]{
    \centering
    \includegraphics[width=0.24\textwidth]{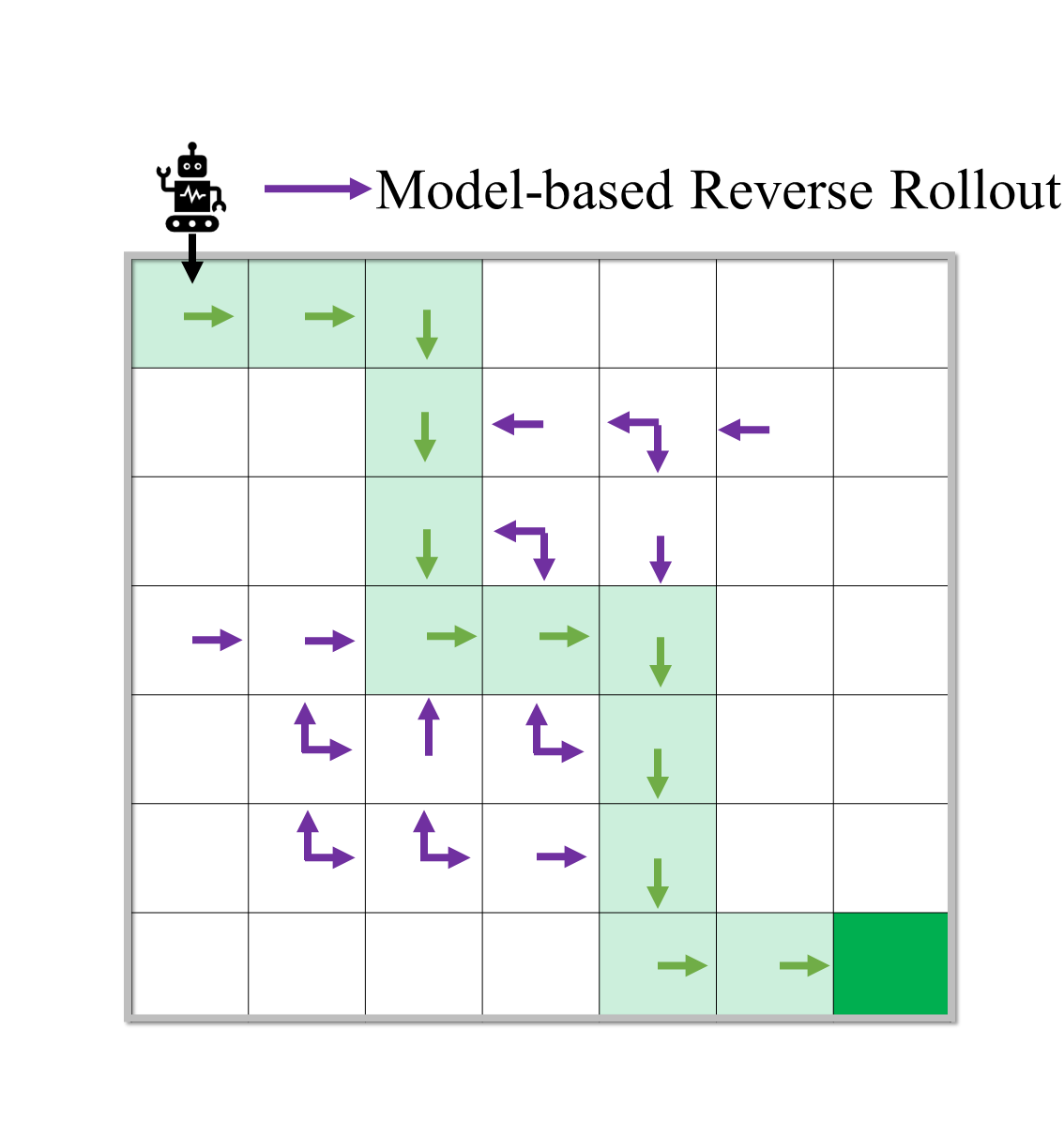}
    \label{motivation_our}
  }
  \label{motivation}
  \caption{The illustration of the main challenges of offline imitation learning and our main idea. (a): The expert data is limited. (b): The preferred action on the expert-unobserved states are uncertain. (c): Model-based forward rollout on expert-unobserved states is difficult to exploit. (d): In this work, we propose utilizing reverse rollout to generate trajectories and leading the agent transit from expert-unobserved states to expert-observed states.}
\end{figure*}

\paragraph{Inverse Reinforcement Learning} Inverse Reinforcement Learning~\cite{DBLP:conf/icml/NgR00} is another popular branch to implement reinforcement learning without reward supervision. 
It involves iteratively learning a reward function and policy~\cite{DBLP:conf/aaai/ZiebartMBD08, DBLP:journals/ai/AroraD21}. However, this requires a potentially large number of online interactions during training, which can result in poor sample efficiency. Recently, offline inverse reinforcement learning has been proposed to eliminate the online interactions and learn from the offline demonstrations~\cite{jarboui2021offline, OTIL}. 
\citet{jarboui2021offline} build a discriminator between expert demonstrations and offline demonstrations 
to serve as the reward function $\hat{r}$, which takes a similar underlying idea like discriminator-based imitation learning, that is, the state-action pairs $(s,a)$ which resemble expert data receive higher rewards. \citet{OTIL} use optimal transport to find an alignment with the minimal wasserstein distance between unlabeled trajectories and expert demonstrations. 
\citet{CLARE} integrates the dynamics conservatism and expert similarity to estimates rewards for the model-based augmented demonstrations, encouraging reinforcement learning to keep the agent in the corresponding area~\cite{CLARE}. 

In the reinforcement learning, there also are some exploration on the methods with reverse dynamics model. This idea is firstly raised in the literature of online reinforcement learning~\cite{holyoak1999bidirectional, goyal2018recall, DBLP:conf/icml/Lai00020, DBLP:conf/icml/LeeSLLS20}. 
\citet{DBLP:conf/icml/Lai00020} utilizes a backward model to reduce the reliance on accuracy in forward model and improve the prediction of the trajectories. 
\citet{ROMI} show the reverse imagination inherently incorporates the conservatism into rollout trajectories in the offline reinforcement learning context. 
The key difference from RL is that in IL the reward function is not available. This determines that rollout samples do not have a reward label. When there is a reward label, RL methods could directly use augmented samples to guide learning. However, in IL, since there is no reward label, existing model-based IL methods try to annotate high rewards on rollout samples similar to expert samples by measuring similarity, which ultimately makes the agent stay around the expert samples (as shown in Figure~\ref{covariate_shift}). 
To our best knowledge, SDA is the first offline IL method with reverse dynamics model to reduce the covariate shift with data augmentation. It reveals that reverse rollout is more easily exploited compared to forward rollout, that is, compared to forward rollout which can only utilize data around the expert data, the reverse rollout is not similar to expert data can also be efficiently utilized. Therefore, it provides a new perspective to the IL community compared to reverse-based RL methods.

\section{The Proposed Method} 
In the offline imitation learning setting, the agent can only access limited expert demonstrations. There are three challenges in this setting. (i) The expert demonstrations has limited samples and generalization over the given data is required. (ii) Without the reward supervision, what action an agent should take when outside the expert-observed states is unknown. (iii) The model-based rollout encounters a lower efficiency of data augmentation utilization when encountering expert-unobserved states, because it doesn't know the preferred action on them. 
Thus it is crucial to augment diverse data on expert-unobserved states while providing preferred action at the same time. 
The Figure~\ref{motivation_our} has demonstrated our main idea. That is, we would like to build the reverse models and generate the trajectories whose endpoints are in the expert-observed states where the agent can perform well, leading the agent from expert-unobserved states to these endpoints, thereby guiding for improving policy's long-term return on these expert-unobserved states. 
In this section, we will introduce the Self-paced Reverse Augmentation (SRA) framework, which encourages diverse data augmentation on expert-unobserved states 
for offline imitation learning.

\subsection{Problem Formulation}
In this work, we consider the infinite Markov Decision Process (MDP) setting~\cite{sutton1998introduction}, denoted as $\mathcal{M} = \{ \mathcal{S}, \mathcal{A}, T, r, d_0, \gamma \}$. Here, $\mathcal{S}$ is the state space, $\mathcal{A}$ is the action space, $T: \mathcal{S} \times \mathcal{A} \rightarrow \Delta(\mathcal{S}) $ is the transition probability of $\mathcal{M}$, $r: \mathcal{S} \times \mathcal{A} \rightarrow [0, 1]$ is the reward function, 
$d_0: \mathcal{S} \rightarrow \Delta(S)$ is the initial state distribution and $\gamma \in (0, 1)$ is the discount factor.
The decision-making process occurs sequentially. At time $t$, the agent observes a state $s_t \in S$ and takes an action $a_t$, following the conditional probability $\pi(a_t|s_t)$. The agent then receives a reward $r(s_t,a_t)$ from the environment, and a new state $s_{t+1}$ appears based on the transition probability $T(s_{t+1}| s_t, a_t)$ of $\mathcal{M}$. The goal of sequential decision-making is to maximize: 
\begin{equation*}
J(\pi)=\mathbb{E}_{s_0\sim d_0, s_{t+1}\sim  T(\cdot|s_t, \pi(s_t)) } \left[ \sum_{t=0}^\infty \gamma^t r (s_t , \pi(s_t)) \right]. 
\end{equation*}

In the offline imitation learning setting, there are limited expert demonstrations with trajectories:
\begin{equation*}
    \begin{aligned}    
  & D^E = \{ (s_0,a_0,s_1,a_1, \dots, s_h, a_h)|  \\
  & s_0 \sim d_0, a_t \sim  \pi^E(\cdot|s_t), s_{t+1} \sim T(\cdot|s_t,a_t), \forall t\in [h]\}. 
    \end{aligned}
\end{equation*}
where $h$ is the length of each trajectory following the expert policy. 
There are also some logged experiences collected by the arbitrary behavior policies $\pi'$ which is much cheap to obtain: 
\begin{equation*}
    \begin{aligned}   
        & D^S = \{ (s_0,a_0,s_1,a_1, \dots, s_h, a_h)| \\
        & s_0 \sim d_0, a_t \sim  \pi'(\cdot|s_t), s_{t+1} \sim T(\cdot|s_t,a_t), \forall t \in [h]\}.
    \end{aligned}
\end{equation*}
In the literature~\cite{d4rl}, the offline policy $\pi'$ can come from various task objectives, suboptimal policies, or even random exploration, which are different from the expert policy and thus challenging to utilize in the imitation learning. 

Model-based offline imitation learning methods typically utilize the union dataset $D^E \cup D^S$ to train a forward dynamic model $\hat{T}$, approximating the ground-truth transition $T$. And the obtained $\hat{T}$ will be used to rollout augmented trajectories with the learned policy $\pi$. In this work, we find the reverse dynamic model $\hat{T}_r$ could generate trajectories form expert-unobserved states to the expert-observed states, which is more efficient for offline imitation learning. 
In the following subsections, we will first introduce the establishment of the reverse models and then the process of data augmentation. 

\begin{table}[t]
  \caption{Summary of Notations.}
  \label{notation}
  \begin{tabular}{cc}
    \toprule
      Notation &  Meaning \\
    \midrule
     $\mathcal{S}$ & State space \\
     $\mathcal{A}$ & Action space \\    
     $\mathcal{T}$ & Transition probability \\
     $D^E$ &  Expert dataset \\
     $D^S$ &  Supplementary dataset \\
     $\gamma$ & Discount factor \\
     $\pi$  & Learned policy \\
     $J(\pi)$ & Expected cumulative return\\
     $\hat{T}_r$ & Reverse dynamic model \\
     $\pi_r$ & Reverse behavioral policy \\
     $\text{Conf}_\pi(s)$ & State-wise policy confidence \\
     $D^A$ & Model-based augmented dataset \\
     $D^R$ & Re-sampled augmented dataset \\
    \bottomrule
  \end{tabular}
\end{table}

\subsection{Reverse Models} 
  In the subsection, we introduce the reverse models, includes reverse dynamic model $\hat{T}_r: \mathcal{S} \times \mathcal{A} \rightarrow \Delta(\mathcal{S})$ and reverse behavior policy $\pi_r:\mathcal{S}\rightarrow \Delta(\mathcal{A})$. The reverse dynamic model want to find which state $s_t$ could lead to states $s_{t+1}$ via action $a_t$, and could be learned from the samples in the union dataset, $(s_t,a_t,s_{t+1})\sim D^E \cup D^S$: 
  \begin{equation}
    \max_{\hat{T}_r} \sum_{(s_t,a_t,s_{t+1})} \log \hat{T}_r(s_t|, s_{t+1},a_t).
    \label{rt}
  \end{equation}
  For the reverse behavior policy, we hope it can help us explore different previous $s_t$, conditioned on $s_{t+1}$. Following the~\cite{DBLP:conf/nips/SohnLY15, DBLP:conf/icml/FujimotoMP19, ROMI}, we build a conditional variational auto-encoder, which contains encoder $\pi_r^e$ and decoder $\pi_r^d$, to approximate the $p(a_t|s_{t+1})$, which could be optimized via maximizing the variational lower bound: 
  \begin{equation*}
    \log \pi_r(a|s) \geq \mathbb{E}_{z\sim \pi_r^e(\cdot|s,a)} \log \pi_r^d(a|z,s) - \text{KL} (\pi_r^e(z|s,a) || p(z|s))
  \end{equation*}
  where $z$ is from the encoded gaussian distribution $z\sim \pi_r^e(s_{t+1},a_t)$ and action $a_t$ could be obtained via the decoder $\pi_r^d(s_{t+1},z)$ with the inputs $s_{t+1}$ and $z$. 
  In the training phase, 
  we simplify the conditional distribution $p(z|s)$ into a normal distribution $\mathcal{N}(0,\mathbf{I})$ and then train the conditional variational auto-encoder through gradient descent, maximizing the above lower bound. 
  In the inference phase, we sample the latent factor $z$ from the multivariate normal distribution $z\sim \mathcal{N}(0,\mathbf{I})$, and then utilize the decoder to produce the conditional actions $p(a_t|s_{t+1})$. 
  Note that the conditional variational auto-encoder $\pi_r$ recover the distribution over the union dataset. 
  If we need to explore extremely diverse states, we could directly sample the $a_t$ from the action space $a_t \sim \text{Uniform}(\mathcal{A})$ with a uniformly random reverse policy $\pi_r$. 
  
\subsection{Self-Paced Augmentation} 

Given the reverse models $\hat{T}_r, \pi_r$ and the target states set $G$, we can perform model-based reverse data augmentation and produce the augmented dataset $D^A$. 
\begin{equation}
  \begin{aligned}
      & D^A = \{s_{-h'}, a_{-h'}, s_{-h'+1}, a_{-h'+1}, \dots, s_{-2},a_{-2},s_{-1}, a_{-1}, s_{0}  |\\
       & s_0\sim G,
        a_{i}\sim \pi_r(s_{i+1}), s_i\sim \hat{T}_r(s_{i+1},a_{i}) \forall i\in [-h', -1] \}
    \label{reverse_rollout}
  \end{aligned}
\end{equation} 
Inspired by self-paced learning~\cite{DBLP:conf/nips/KumarPK10, DBLP:conf/nips/JiangMYLSH14, DBLP:journals/isci/MengZJ17}, we would like to cautiously and adaptively prioritize learning simple, reliable examples, and then gradually transition to learning difficult examples. 
Instead of directly using expert data as the target states $G$ and generating a fixed augmented dataset, we judge whether the policy has well learned the behavior of the state based on its self-confidence, gradually expanding the $G$ to generate more diverse data. 
Formally, we take the self-confidence as the measure of an agent's state-wise belief degree. 
\begin{equation}
  \text{Conf}_\pi(s) = \pi(\mathbb{E}[\pi(a|s)]|s)
\end{equation}
In areas that have been fully learned, such as expert-observed states, the policy has a high conditional probability 
for the expected action $\mathbb{E}[\pi(a|s)]$, and therefore has a high $\text{Conf}_\pi(s)$. 
Therefore, we construct the set of target points $G$ with the $\text{Conf}_\pi(s)$:  
\begin{equation}
  G = \{s| \text{Conf}_\pi(s) \geq \mathbb{E}_{s'\sim D^E} \text{Conf}_\pi(s') \} \cup \{s'|s' \in D^E\} 
  \label{goal_states}
\end{equation}

Moreover, in expert-unobserved states, the policy has lower $\text{Conf}_\pi(s)$ in determining which action is preferred. 
To explore more expert-unobserved states and mitigate the covariate shift, we further design a data selection mechanism that over-samples the under-exploring states to enhance the agent on them. 
Specifically, we over-sample the states which has lower $\text{Conf}_\pi(s)$, the state-wise sampling probability is: 
\begin{equation}
  p(s) = 1 / \text{Conf}_\pi(s)
  \label{sample_weight} 
\end{equation}

With the weighted sampling mechanism, we could obtain a re-sampled dataset $D^R$, encouraging the more diverse data with under-exploring states. 

\subsection{Algorithm Formalization}
\label{subsection34}
Based on the learned reverse models and self-paced augmentation, we could explore the expert-unobserved states and collect the dataset $D^R$. 
Since our reverse augmentation is agnostic to the specific learning of policy $\pi$, the proposed data augmentation could be combined with any model-free offline imitation learning methods. 
In order to fully utilize augmented data and lead the agent to transit from expert-unobserved states to expert-observed states, we consider utilizing the offline reinforcement learning methods which include dynamic programming objective, such as IQL~\cite{IQL} and TD3BC~\cite{TD3BC}. To keep the simplicity of our proposal, we do not add the additional reward learning module from the inverse reinforcement learning methods, but label the expert samples reward as 1, the rest are all 0, which is known as unlabeled data sharing~\cite{UDS, DBLP:conf/iclr/0006YZZ23,BCDP}. It has shown advantages in utilizing supplementary sub-optimal data, and is sufficient enough to implement our objective, leading the agent to expert-observed states that have the higher rewards. 
The overall framework are shown in Algorithm~\ref{algo}.

\begin{algorithm}[tb]
  \caption{Offline IL with Self-Paced Reverse Augmentation}
  \label{algo}
  \begin{algorithmic}[1] 
  \REQUIRE Expert dataset $D^E$, supplementary dataset $D^S$, length of reverse rollout $h'$, The number of iterations $N_D$, $N_R$, $N_P$. 
  \FOR{$t=1$ to $N_D$}
    \STATE Update the reverse dynamic model $\hat{T}_r$ with gradient descent. 
  \ENDFOR
  \FOR{$t=1$ to $N_R$}
    \STATE Update the reverse behavior policy $\pi_r$ with gradient descent. 
  \ENDFOR
  \FOR{$t=1$ to $N_P$}
    \STATE Collect the goals $G$ via Eq.~\ref{goal_states}. 
    \STATE Rollout the reverse trajectories $D^A$ with $G, \hat{T}_r, \pi_r$, via Eq.~\ref{reverse_rollout}. 
    \STATE Re-sample the augmented samples to obtain $D^R$, with the sampling weights (Eq.~\ref{sample_weight}).
    \STATE Get the union dataset $D^U = D^E \cup D^S \cup D^R$ 
    \STATE Train the policy $\pi$ with model-free reinforcement learning methods on the union dataset.  
  \ENDFOR
  \STATE \textbf{return} Policy $\pi$.
  \end{algorithmic}
\end{algorithm}

\section{Empirical Study}




\subsection{Experimental Setting}


We evaluate on a wide range of domains in the D4RL benchmark~\cite{d4rl}, includes navigation and locomotion. 

\textit{Navigation.} We conduct the experiments on the \emph{Maze2D} environments to evaluate the policy. The \emph{Maze2D} domain requires an agent to navigate in the maze to reach a fixed target goal and stay there. The D4RL benchmark provides three maze layouts (i.e., umaze, medium, and large) and two rewards types (i.e., sparse and dense reward singals) in this domain. We employs 5 expert trajectories as the expert data which follows a path planner~\cite{d4rl}. We consider the offline data as the logged experience with random goals (umaze-v1, medium-v1, and large-v1.), which provided by the D4RL benchmark. They evaluate the ability of offline imitation learning methods to use data from related tasks. 

\textit{Locomotion.}
We conduct the experiments on the \emph{Gym-MuJoCo} environments to evaluate the policy. It consists four different environments (i.e., hopper, walker2d, halfcheetah and ant). We employs 5 expert trajectories from the ``-expert" datasets as the expert data. We also consider offline supplementary from sub-optimal policy (hopper-medium-v2, hopper-medium-expert-v2, etc.). The ``medium" dataset contains logged experiences from an early-stopped SAC policy, while the ``medium-expert" dataset comes from a mixture of  early-stopped SAC policy and well-learned SAC policy. 


\textbf{Competing Baselines:} We compare SRA with the well-validated baselines and state-of-the-art offline imitation learning methods: 
  \begin{itemize}
    \item \textbf{BC-exp}: Behavior cloning on expert data $D^E$. The $D^E$ contains high-quality demonstrations but with limited quantity, and thus causes serious generalization problem. 
    \item \textbf{DemoDICE}~\citep{DemoDICE}: DemoDICE approximates the state-action distribution $d^\pi(s,a)$ to both expert data with offline data, treating offline data as a regularization, i.e., $\min_\pi KL(d^\pi||D^E)$ $+ \alpha KL(d^\pi||d^o)$ with the expectation of further improving performance based on expert data.
    \item \textbf{DWBC}~\citep{DWBC}: DWBC regards the offline data as a mixture of expert-similar trajectories and low-quality trajectories, and apply positive-unlabeled learning to build a discriminator. The discriminator will evaluate unlabeled trajectories and provide an expert-similarity score, followed by a weighted behavior cloning.
    \item \textbf{OTIL}~\citep{OTIL}: OTIL uses optimal transport to label the rewards of offline trajectories based on their Wasserstein distance from expert trajectories. It then employs offline reinforcement learning algorithms to train an agent on the labeled dataset. We implement offline RL using IQL~\citep{IQL}, as described in the original paper. 
    \item \textbf{MILO}~\cite{MILO}: MILO attempts to mitigate the covariate shift, though conducting the adversarial imitation learning with the learned dynamic model.   
    \item \textbf{CLARE}~\cite{CLARE} CLARE integrates the dynamics conservatism and expert similarity to estimates rewards for the model-based augmented demonstrations. The the policy is learned with the reward function and subsequent offline reinforcement learning methods. We chosen the well-validated offline RL algorithm, IQL, to implement the CLARE. 
    \item \textbf{UDS}~\citep{UDS}: UDS 
    annotates 
    the unlabeled datasets with reward 0 (minimum rewards), and uses offline RL 
    to train the agent on the merged dataset. 
    Compared to their setting has a high-quality labeled dataset, our expert dataset does not have a ground-truth reward label. Instead, we label them with 1 (maximum rewards) to implement offline RL. 
    We also choose the IQL as the specific algorithm for consistence. 
    \item \textbf{ROMI}~\cite{ROMI} is also a data augmentation method based on reverse models, which generate conservative behaviors. Despite it being a method proposed for offline RL, we also regard it as one of our ablation studies here, and have made comparisons. 
  \end{itemize}

  \begin{table*}[t]
    \caption{The results on D4RL benchmark. All values are normalized to lie between 0 and 100, where 0 corresponds to a random policy and 100 corresponds to an expert. 
    The best result in each setting is in bold.}
    \setlength{\tabcolsep}{0.5mm}{
    \begin{tabular}{c|cccc|ccc|cc}
      \hline\hline
      DataSet & BC-exp & DemoDICE & DWBC & OTIL & CLARE & MILO & ROMI & UDS & SRA \\
      \hline
      maze2d-umaze-sparse-v1 & 100.$\pm$11.6 & 88.7$\pm$10.4 & 25.8$\pm$5.65 & 128.$\pm$8.22  & -3.08$\pm$6.02 & 75.0$\pm$11.2 & 154.$\pm$5.96 & 64.9$\pm$8.51 & \textbf{155.$\pm$6.20}\\
      maze2d-medium-sparse-v1 & 44.6$\pm$11.1 & 15.4$\pm$7.83 & 22.7$\pm$4.77 & 98.2$\pm$11.0 & 33.5$\pm$7.82 & 47.9$\pm$13.5 & 123.$\pm$10.5 & 83.0$\pm$8.84 & \textbf{147.$\pm$5.67} \\
      maze2d-large-sparse-v1 & 15.5$\pm$7.98 & 8.68$\pm$3.55 & 35.1$\pm$4.18 & 129.$\pm$14.4 & 18.6$\pm$9.12 & 51.2$\pm$17.1 & 101.$\pm$20.2 & 108.$\pm$16.7 & \textbf{150.$\pm$14.9}\\
      \hline 
      maze2d-umaze-dense-v1 & 70.6$\pm$9.55 & 69.1$\pm$9.21 & 39.2$\pm$4.77 & 100.$\pm$6.98 & 5.84$\pm$6.61 & 54.9$\pm$6.96 & 111.$\pm$6.23 & 62.3$\pm$6.99 & \textbf{113.$\pm$5.80}\\
      maze2d-medium-dense-v1 & 45.0$\pm$10.2 & 34.3$\pm$7.08 & 39.1$\pm$3.34 & 95.7$\pm$8.66 & 46.3$\pm$7.81 & 44.4$\pm$11.0 & 112.$\pm$9.10 & 87.3$\pm$7.80 & \textbf{138.$\pm$5.29}\\
      maze2d-large-dense-v1 & 18.2$\pm$8.57 & 21.7$\pm$6.30 & 56.1$\pm$5.56 & 120.$\pm$11.0 & 26.5$\pm$8.78 & 40.7$\pm$14.0 & 101.$\pm$16.6 & 109.$\pm$14.4 & \textbf{140.$\pm$11.4}\\
      \hline 
      hopper-medium & 72.9$\pm$5.50 & 54.1$\pm$1.67 & 88.1$\pm$4.71 & 26.2$\pm$2.28 & 82.2$\pm$6.56 & 75.0$\pm$7.46 & 67.3$\pm$4.82 & 59.5$\pm$4.51 & \textbf{90.2$\pm$4.93}\\
      halfcheetah-medium & 13.3$\pm$2.74 & 41.1$\pm$1.00 & 22.5$\pm$3.94 & 38.7$\pm$0.75 & 32.2$\pm$3.14 & 41.9$\pm$0.92 & 43.6$\pm$1.53 & 43.6$\pm$5.15 & \textbf{43.7$\pm$1.72}\\
      walker2d-medium & 99.1$\pm$3.66 & 73.0$\pm$2.09 & 84.8$\pm$5.65 & 86.9$\pm$3.63 & 49.9$\pm$5.37 & 67.9$\pm$3.13 & 96.6$\pm$3.76 & 97.6$\pm$2.85 & \textbf{101.$\pm$3.60}\\
      ant-medium & 51.3$\pm$6.87 & 91.2$\pm$3.79 & 37.5$\pm$5.95 & 72.4$\pm$5.68 & 68.5$\pm$7.35 & 92.0$\pm$3.55 & \textbf{92.7$\pm$6.46} & 87.3$\pm$5.10 & 88.9$\pm$7.18\\
      \hline 
      hopper-medium-expert & 72.9$\pm$5.50 & 98.6$\pm$4.32 & 99.4$\pm$4.43 & 42.5$\pm$3.70 & 93.9$\pm$5.81 & 90.9$\pm$5.42 & 100.$\pm$3.40 & 97.4$\pm$3.35 & \textbf{104.$\pm$3.37}\\
      halfcheetah-medium-expert & 13.3$\pm$2.74 & 48.9$\pm$5.46 & \textbf{82.3$\pm$3.79} & 43.7$\pm$2.76 & 31.4$\pm$5.15 & 44.5$\pm$1.57 & 58.8$\pm$3.29 & 67.1$\pm$2.63 & 63.4$\pm$3.52\\
      walker2d-medium-expert & 99.1$\pm$3.66 & 93.1$\pm$5.49 & \textbf{106.$\pm$1.57} & 82.5$\pm$2.76 & 39.9$\pm$7.66 & 95.4$\pm$3.87 & 103.$\pm$2.12 & 103.$\pm$2.32 & 104.$\pm$4.88\\
      ant-medium-expert & 51.3$\pm$6.87 & 69.8$\pm$7.97 & 58.2$\pm$8.81 & 79.2$\pm$7.40 & 3.61$\pm$2.86 & \textbf{115.$\pm$4.63} & 105.$\pm$6.90 & 92.2$\pm$8.14 & 94.1$\pm$7.86\\
    \hline\hline
  \end{tabular}}
  \label{main_exp}
  \end{table*}

  For a fair comparison, we use the same actor network as~\cite{DWBC} and same critic networks as~\cite{IQL}. 
  For the learning of dynamic models, we keep the same network architecture and the corresponding learning process as~\cite{ROMI} for all methods which needs forward model or reverse model.  
  Moreover, for methods that require the use of offline reinforcement learning, we consistently choose IQL, including our SRA. 
  We do not put much effort into the hyperparameter tuning. All hyper-parameters are fixed with the original IQL implementation. Regarding reverse augmentation, in the navigation task, we use a uniformly random reverse policy to generate as diverse actions as possible, in the locomotion task, we use a conditional variational auto-encoder to generate reasonable actions. The length of reverse rollout is fixed as 5 across different settings. 
  For all settings, we obtain undiscounted average returns of the policy with 30 evaluations after training. 
  Evaluation results are averaged over three random seeds, and deviation is measured by 95\% bootstrapped confidence interval. 
  
  \subsection{Main Results}
  We evaluate SRA in D4RL benchmark domains with 14 settings and provide the results in the Table~\ref{main_exp}. 
  From the results, we have the following observations and analyses: 
  The unsatisfactory performance of BC-exp baseline indicates the difficulty of imitating learning when the expert data is limited. 
  DemoDICE uses all of the offline data as a regularization for behavior cloning, without considering the possibility that some of the unlabelled data may be not suitable for direct use in behavior cloning, sometimes leading to weaker performance compared to BC-exp. 
  DWBC uses positive-unlabeled learning to identify samples similar to expert behavior, which can filter out the sub-optimal samples and reduce harm from them. Despite this, the performance gain it brings is limited when the overall data quality is not high. 
  Compared with other model-free offline imitation learning methods, the OTIL method performs better overall. One plausible reason is that, although OTIL does not directly focus on the trajectories from expert-unobserved states to expert-observed states, the part of Q-learning in the IQL it uses implicitly does this, which may lead to a satisfactory performance of OTIL. 
  Model-based methods, MILO and CLARE perform better on locomotion tasks, but perform worse on navigation tasks, even weaker than the BC baseline and other model-free methods. This is because covariate shift is more serious in navigation tasks, and the agent needs to make decisions in many expert-unobserved states. Although MILO and CLARE train the dynamic model to augment the dataset, they encourage the policy to make decisions in areas near the expert-observed states, ignoring the exploration of expert-unobserved states. 
  In the navigation tasks, ROMI performs well, which also indicates the potential of using reverse rollout to mitigate covariate shift. Similarly, the dynamic programming process of UDS, as we pointed out, can transfer the agent from expert-unobserved states to expert-observed states, thereby improving the long-term return, and the overall performance is also acceptable. 
  It is worth noting that our method, SRA, achieved the best result in all settings in navigation tasks, which is a domain with serious covariate shift problems. In locomotion tasks, SRA also achieved competitive performance. These phenomena clearly demonstrate the effectiveness of our proposal and indicate it provides a promising way to mitigate the covariate shift of offline imitation learning. 

  \subsection{Concerns on Self-Paced Process}

\begin{figure*}[t]
    \centering
    \includegraphics[width=.9\textwidth]{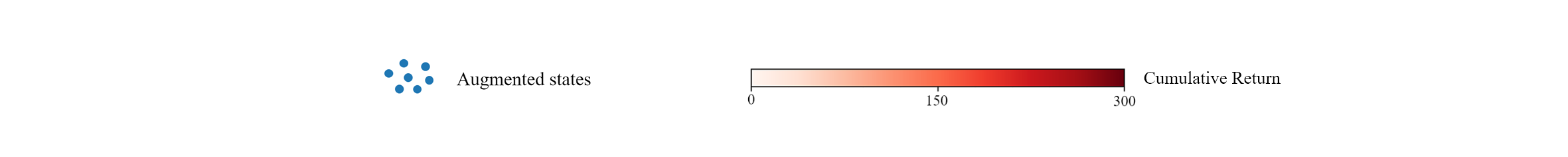}\\
    \subfigure[][State-wise cumulative return of policy $\pi$ after 0, 50000, 100000, and 150000 iterations.]{    
    \includegraphics[width=.24\textwidth]{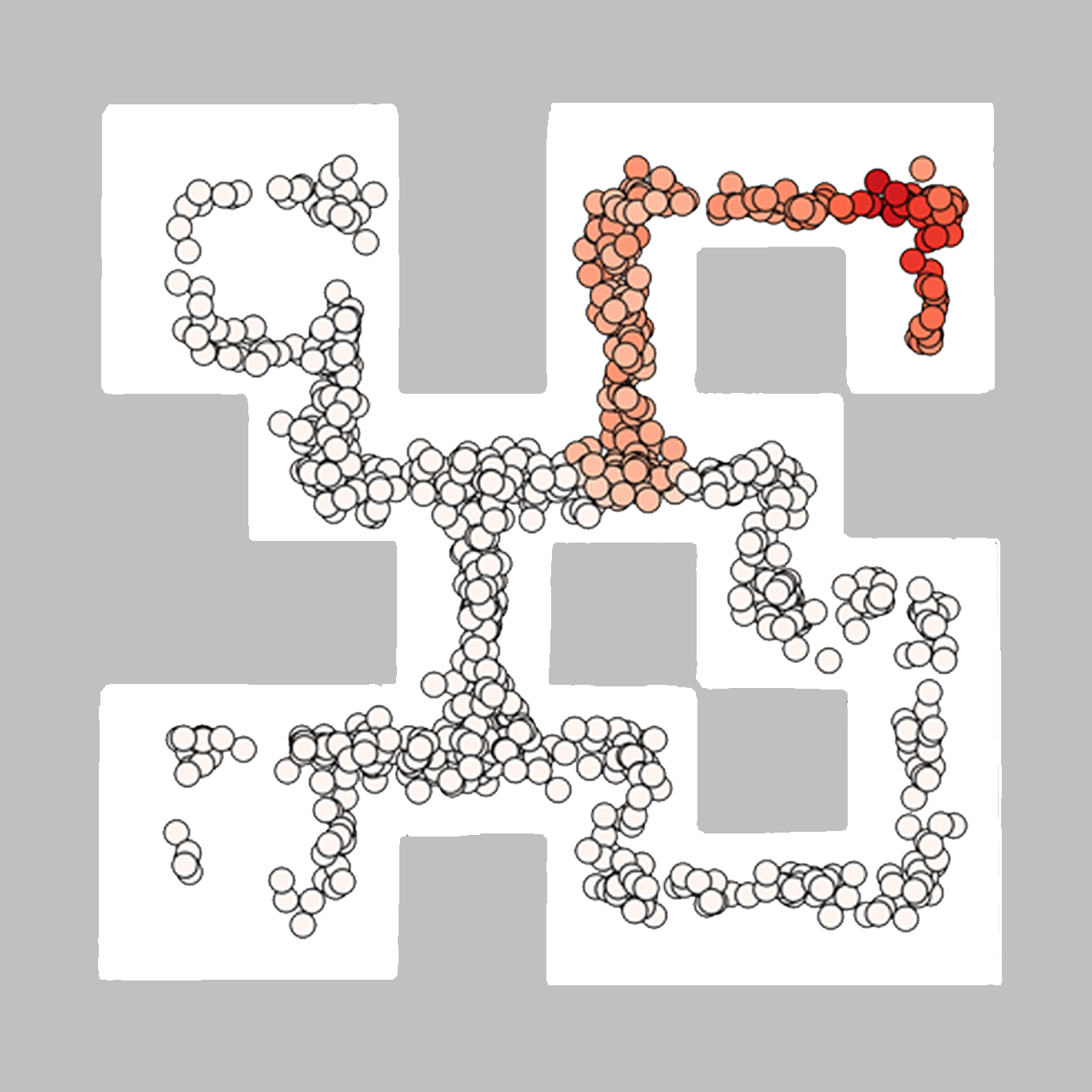}
    \includegraphics[width=.24\textwidth]{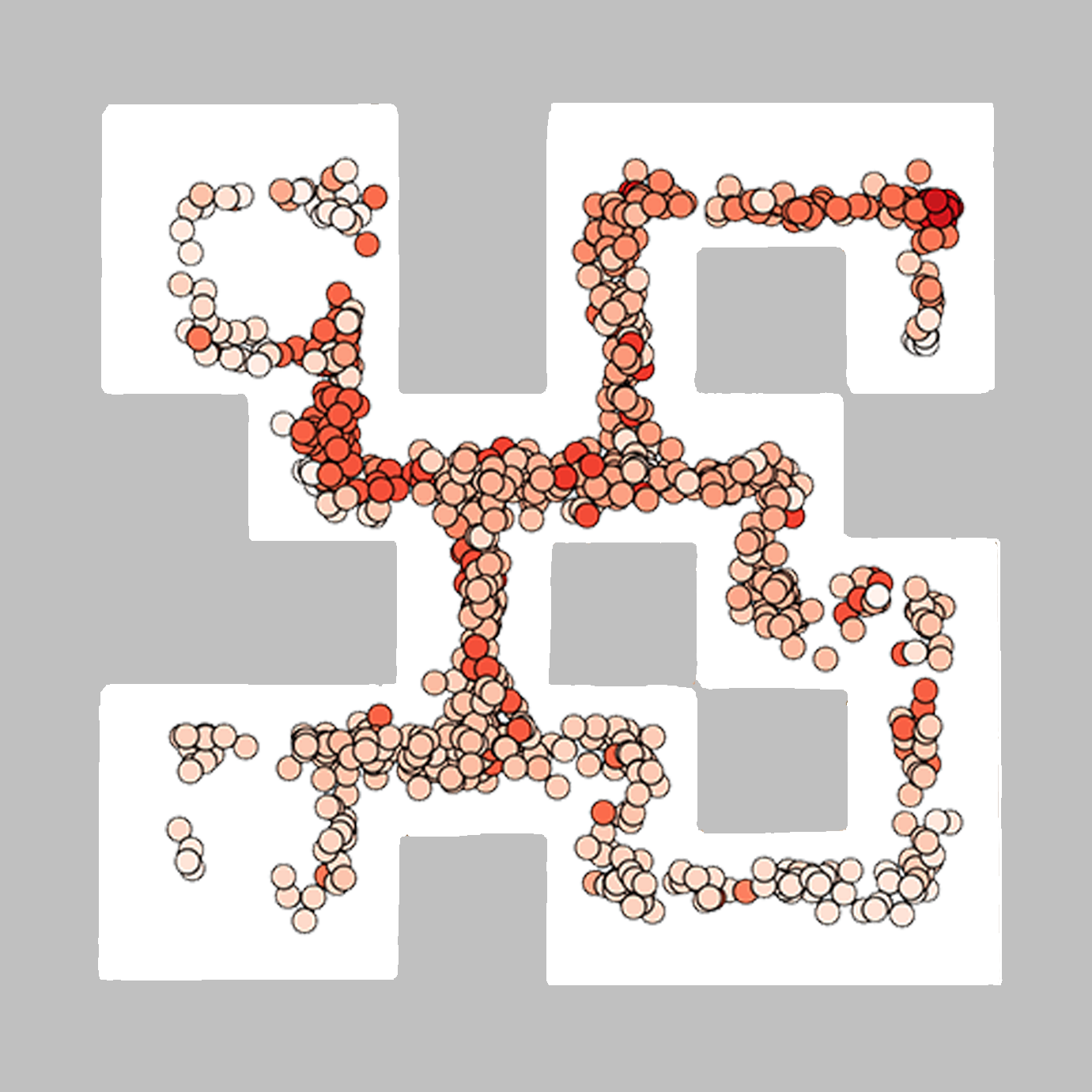}
    \includegraphics[width=.24\textwidth]{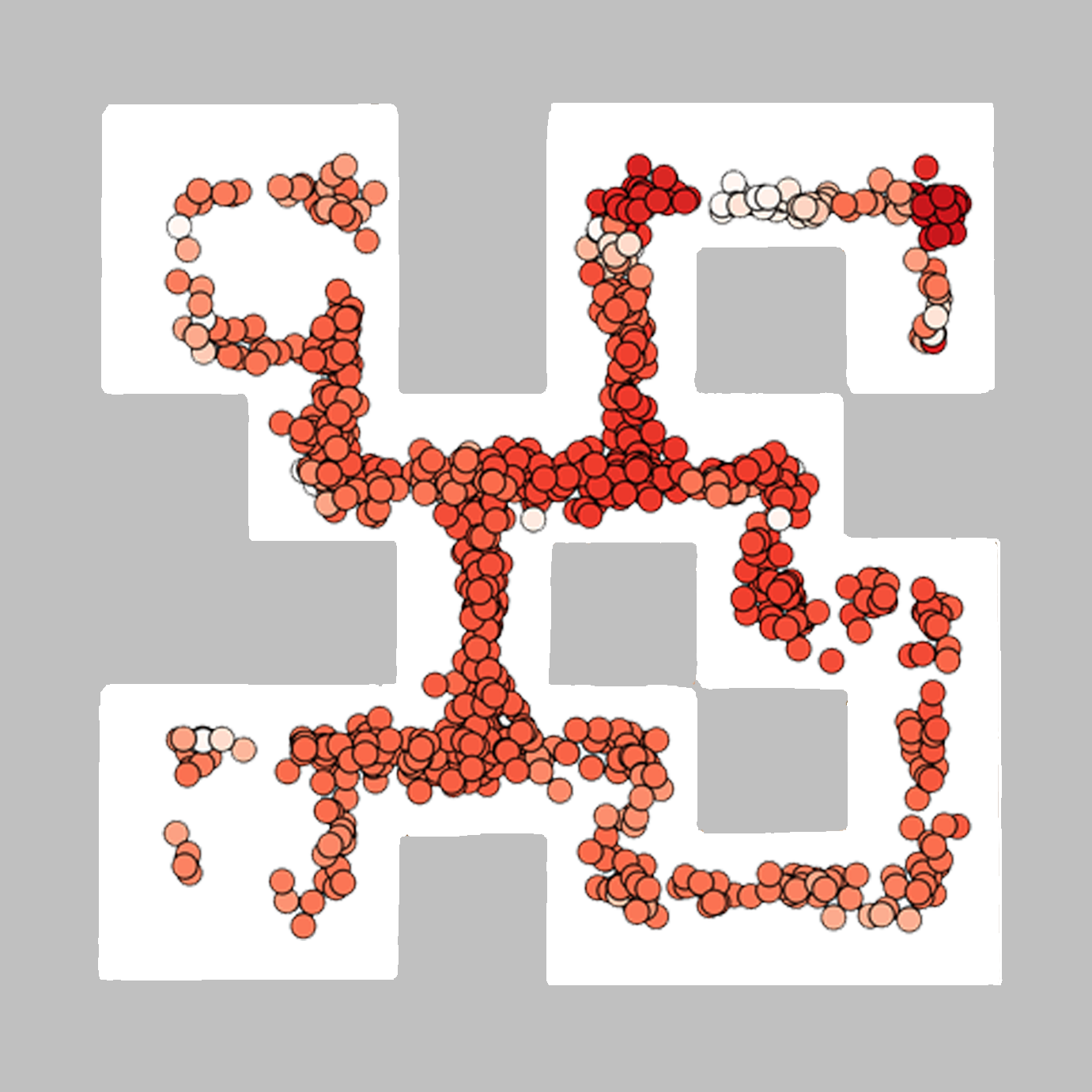}
    \includegraphics[width=.24\textwidth]{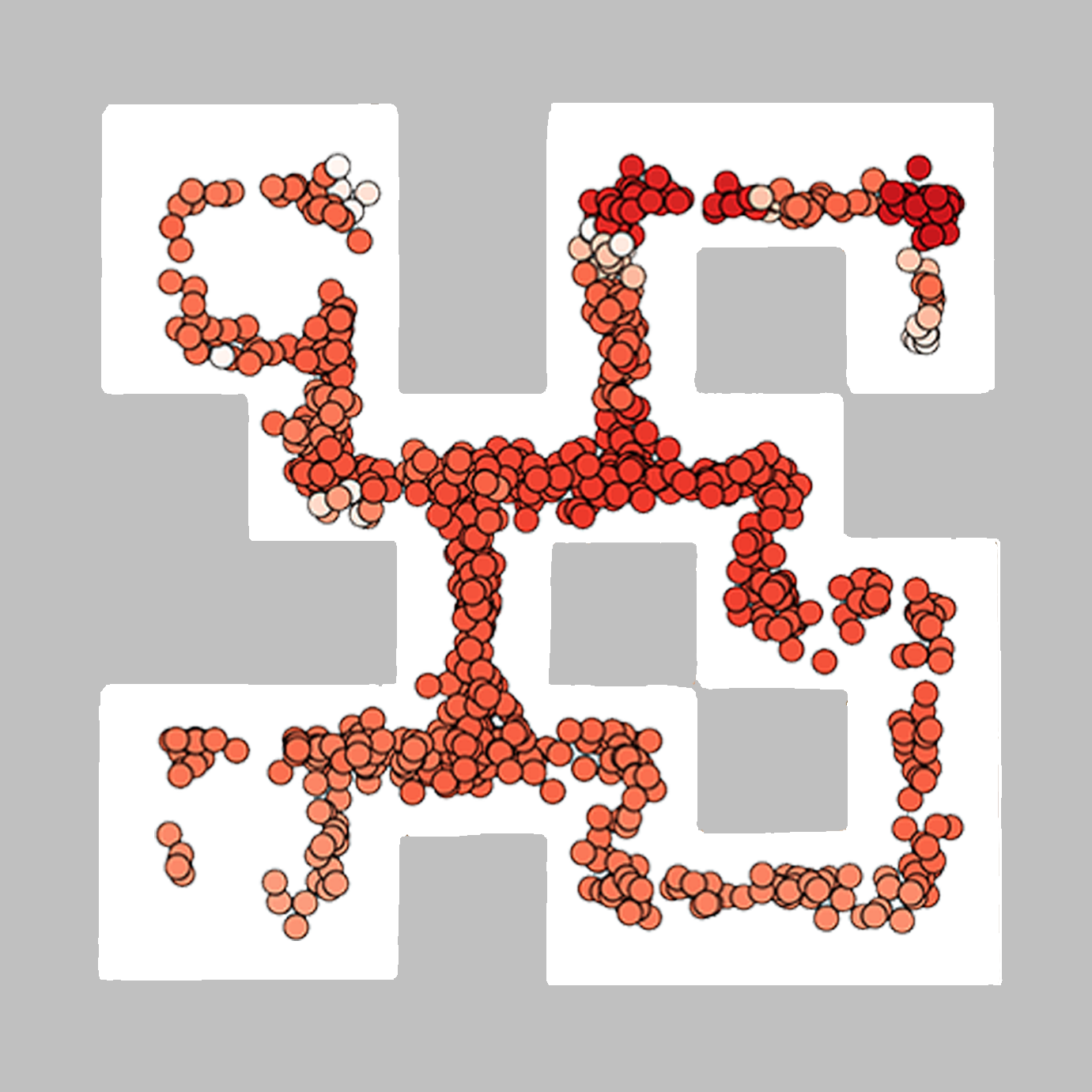}}
    \subfigure[][Re-sampled augmented dataset $D^R$  after 0, 50000, 100000, and 150000 iterations.]{
    \includegraphics[width=.24\textwidth]{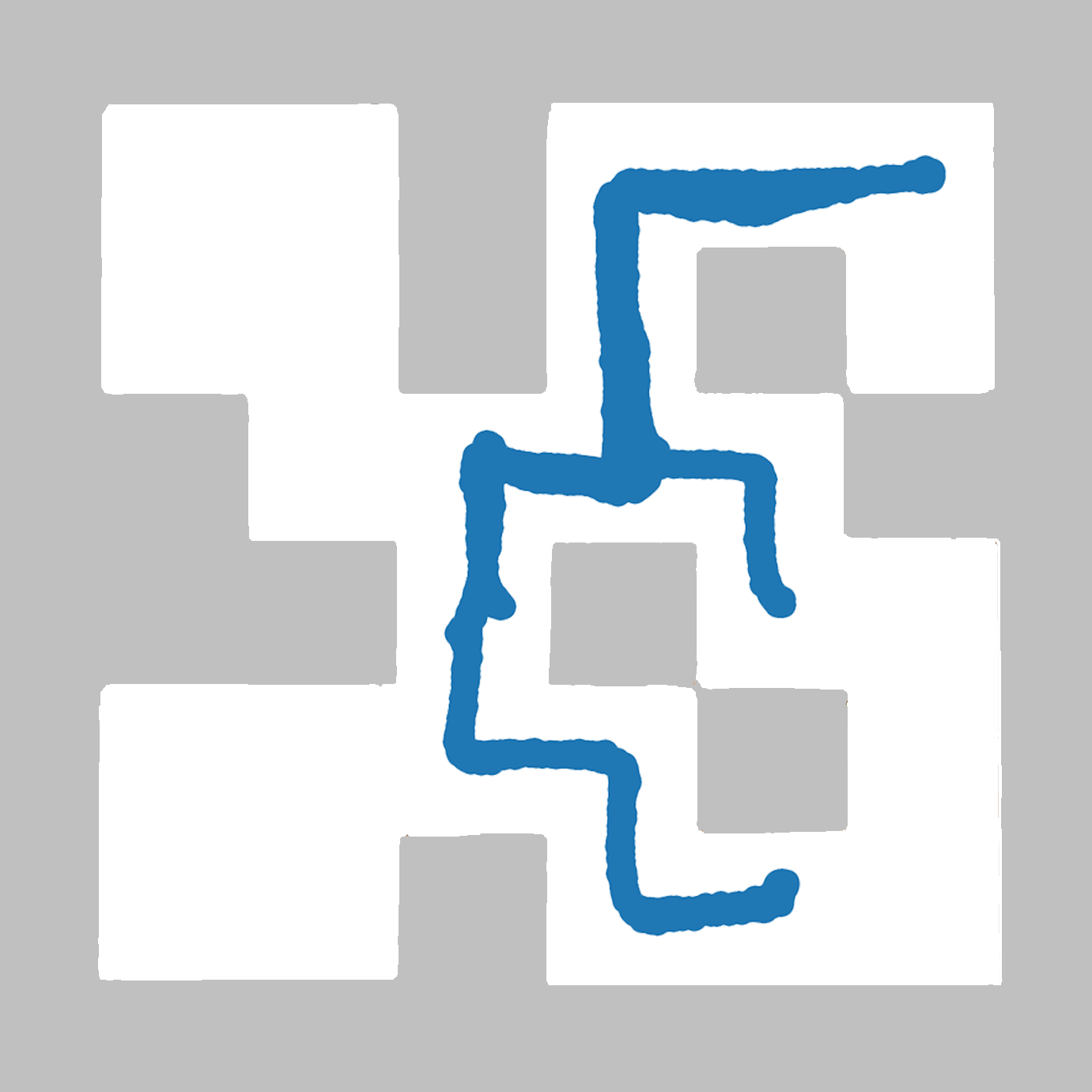}
    \includegraphics[width=.24\textwidth]{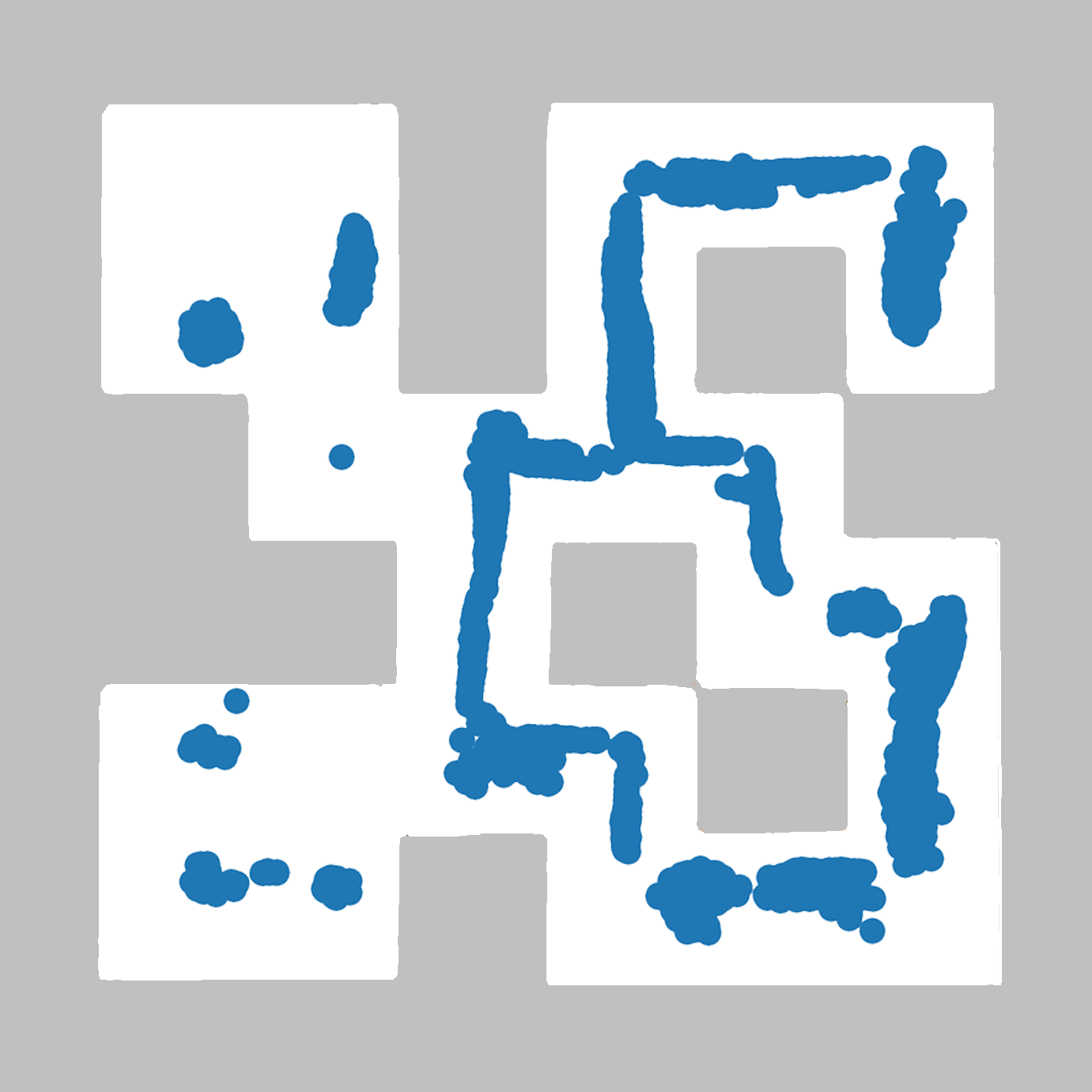}
    \includegraphics[width=.24\textwidth]{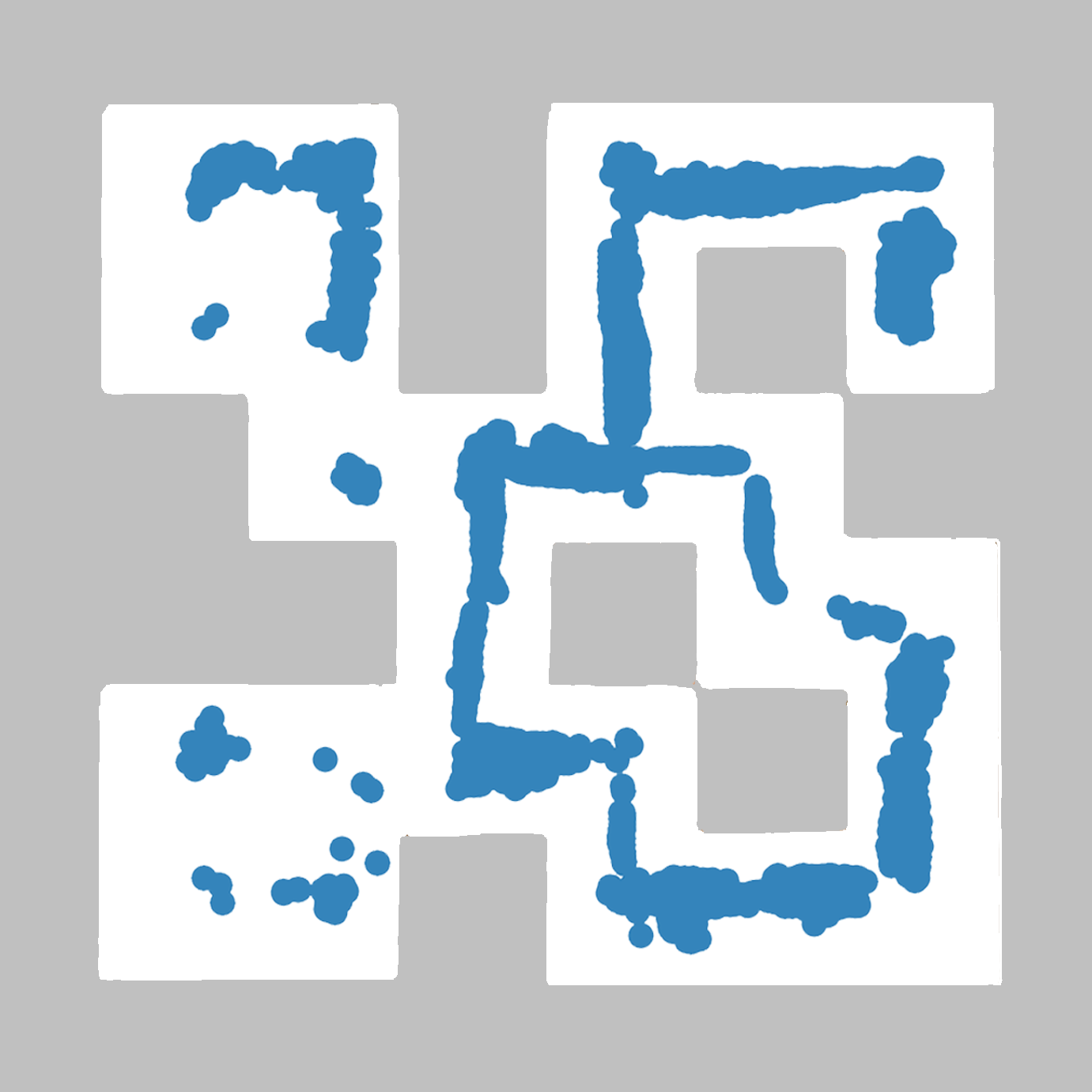}
    \includegraphics[width=.24\textwidth]{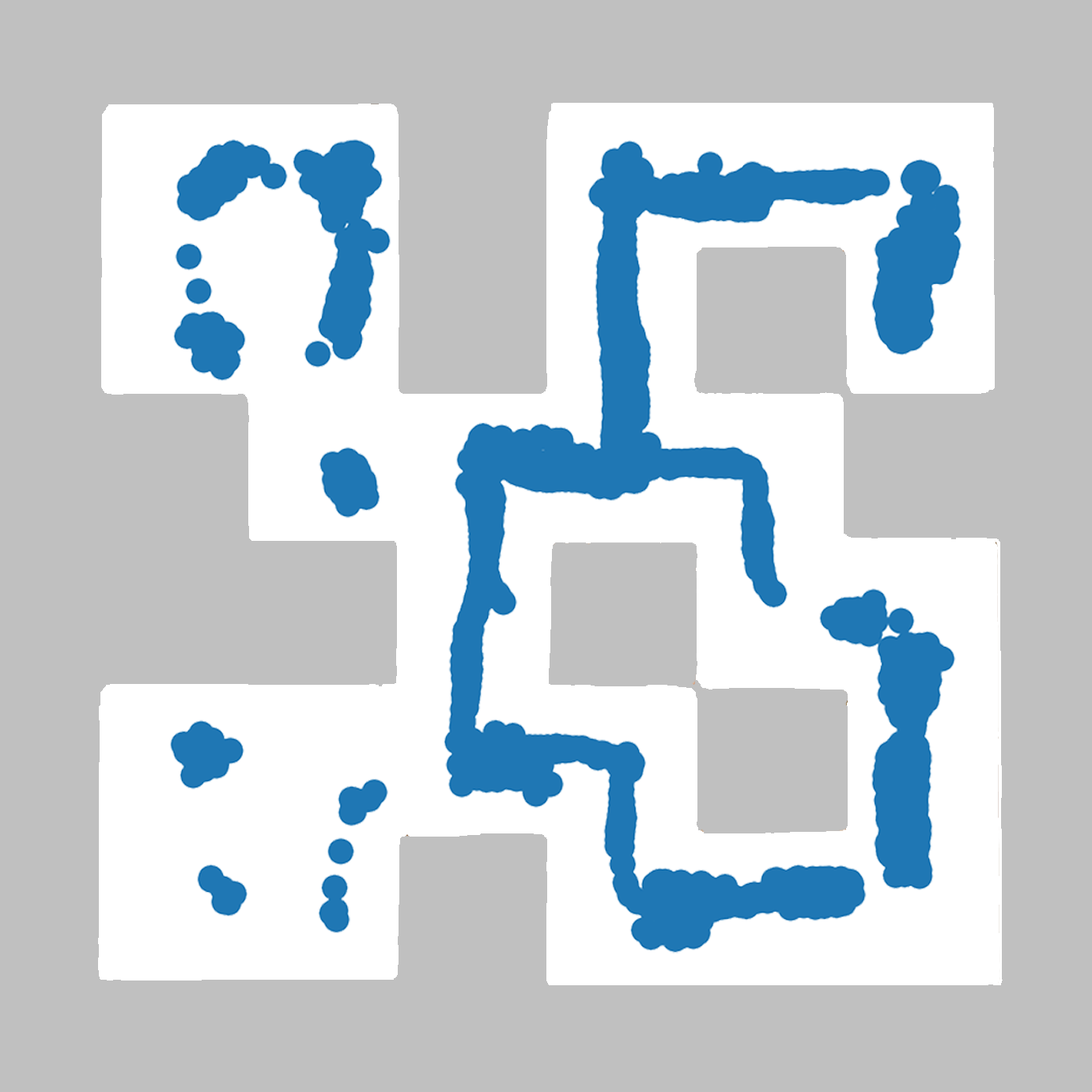}}
    \caption{Visualization during Self-paced Learning Process}
    \label{sp_learning}
  \end{figure*}

  In this work, we adaptively prioritize learning simple, reliable examples, and then gradually transition to exploring diverse expert-unobserved states. To dive deeper into the learning process, we conduct the visualization to demonstrate the learning process in the \textit{Maze2D-Medium} environment. As shown in the Figure~\ref{sp_learning}, we visualize the state-wise cumulative return of policy $\pi$ and the data coverage of corresponding sampled dataset $D^A$. 

  \begin{table}[t]
    \caption{Ablation study with the Self-Paced module.}
    \label{ablation}
    \begin{tabular}{c|c|c}
      \hline\hline
      DataSet & SRA w/o SP & SRA \\
      \hline
      Umaze-sparse & 143. $\pm$ 6.92	& \textbf{155. $\pm$ 6.20} \\
      Medium-sparse & 110. $\pm$ 14.7	& \textbf{147. $\pm$ 5.67} \\
      Large-sparse & 92.0 $\pm$ 21.7 & \textbf{150. $\pm$ 14.9} \\
      \hline 
      Hoppper-medium & 69.2 $\pm$ 5.52 & \textbf{90.2 $\pm$ 4.93} \\
      Halfcheetah-medium & \textbf{44.2 $\pm$ 1.33} & 43.7 $\pm$ 1.72 \\
      Walker2d-medium	& 96.8 $\pm$ 5.41 & \textbf{101. $\pm$ 3.60} \\
      Ant-medium & 87.5 $\pm$ 6.28 & \textbf{88.9 $\pm$ 7.18} \\ 
    \hline\hline
    \end{tabular}
  \end{table}

  \begin{figure*}[t]
    \centering
    \includegraphics[width=.9\textwidth]{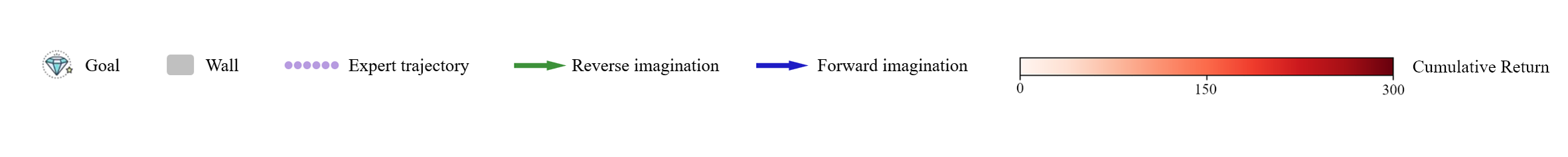}\\
    \includegraphics[width=.24\textwidth]{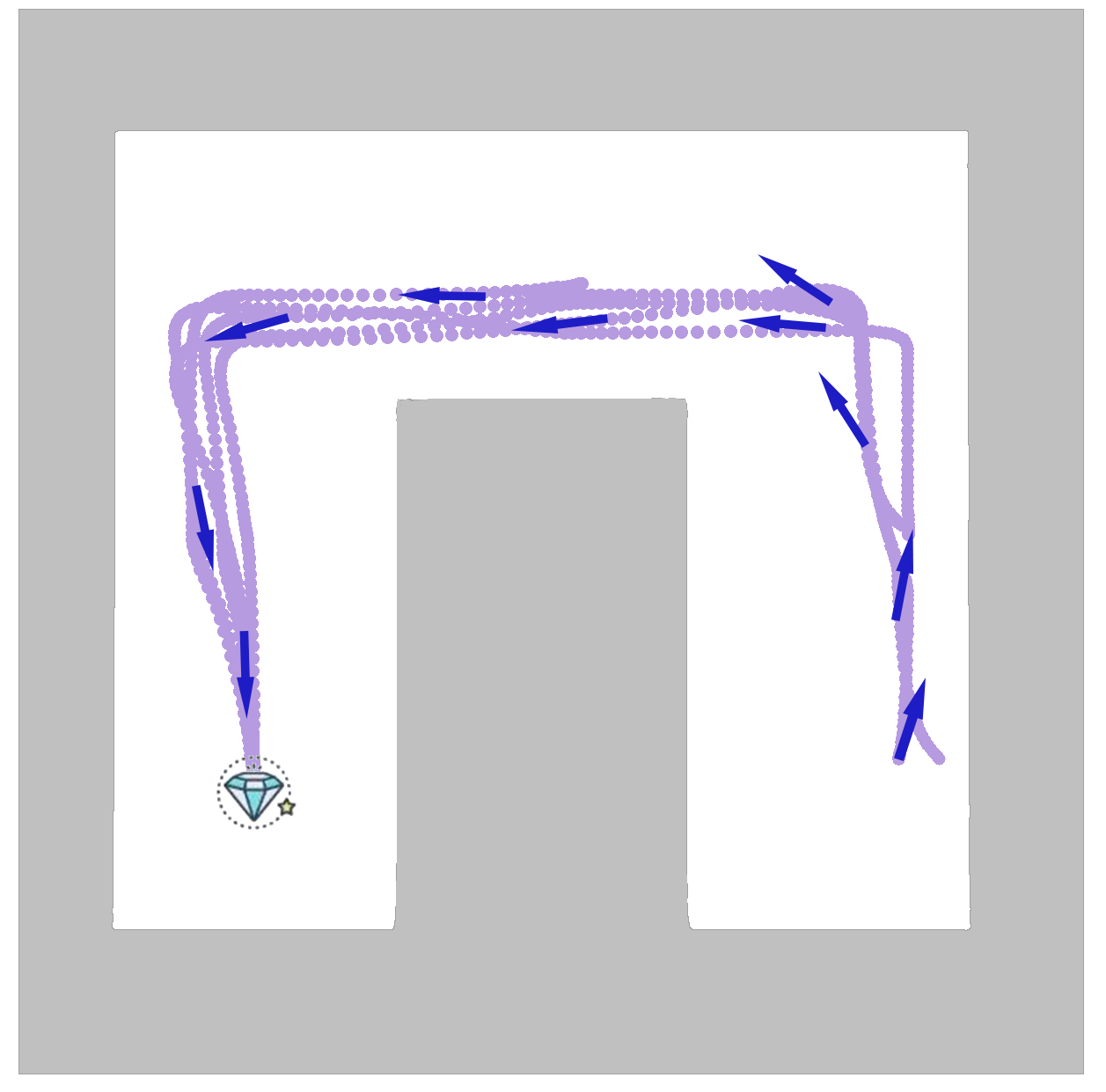}
    \includegraphics[width=.24\textwidth]{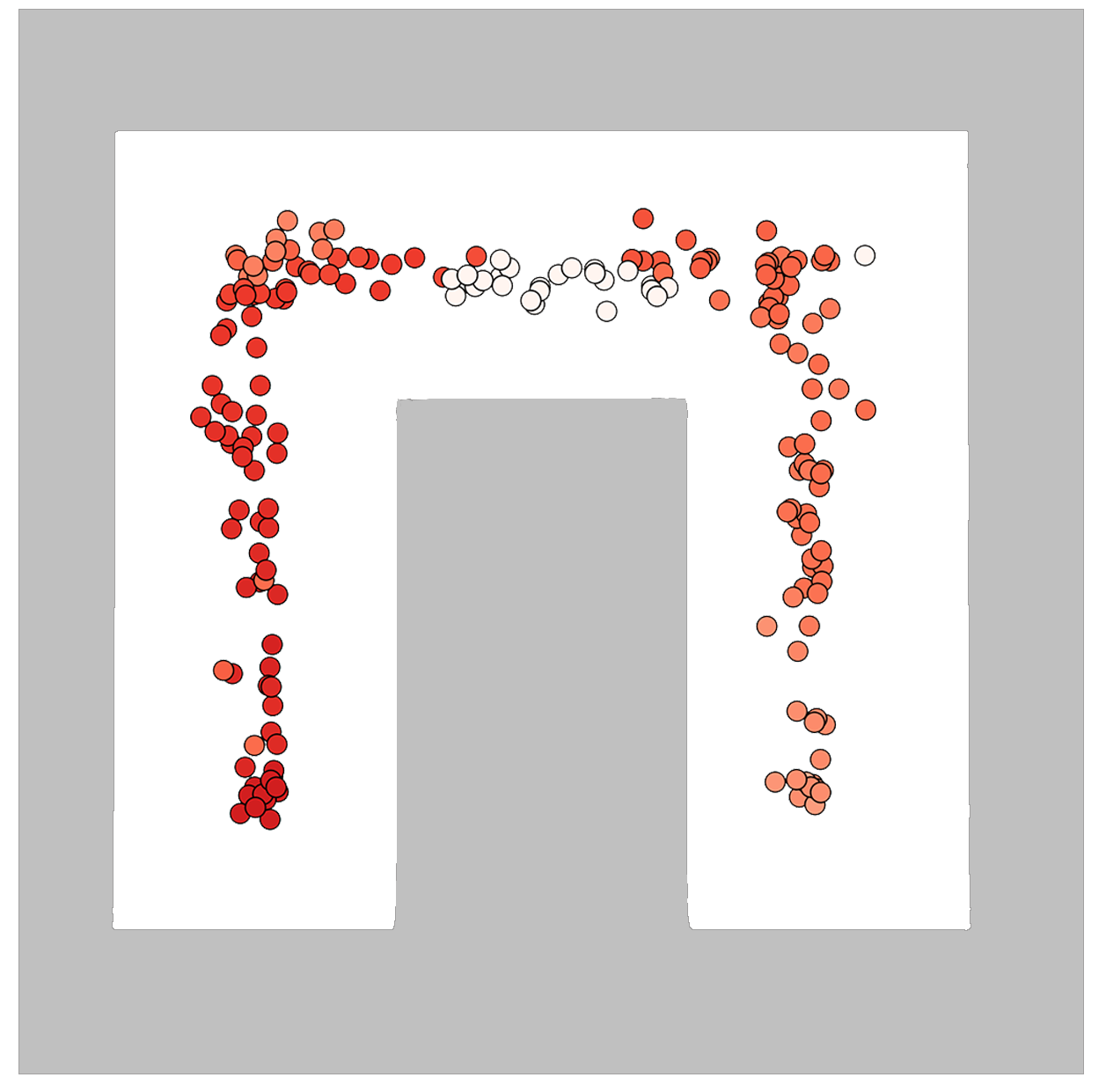}
    \includegraphics[width=.24\textwidth]{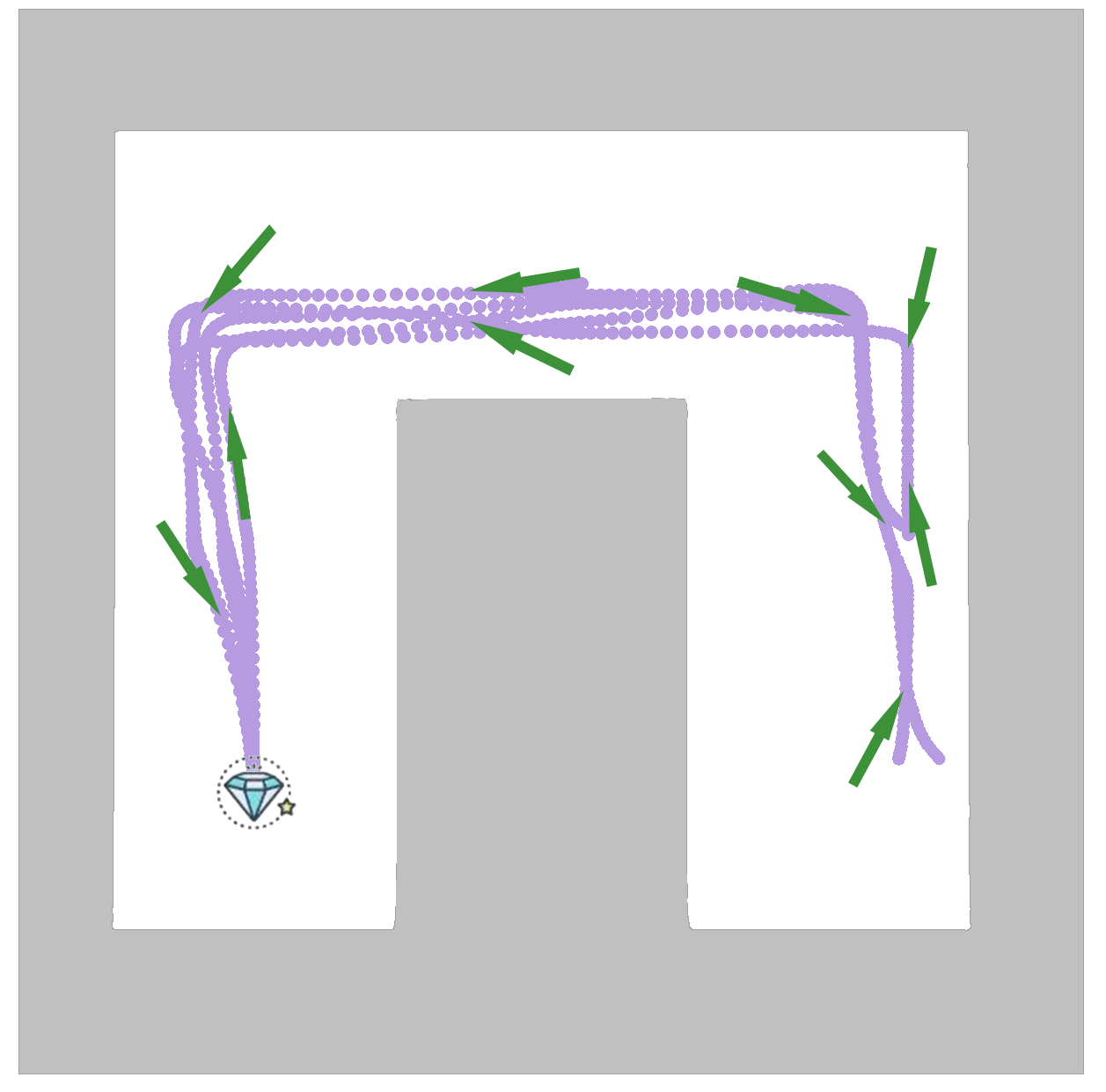}
    \includegraphics[width=.24\textwidth]{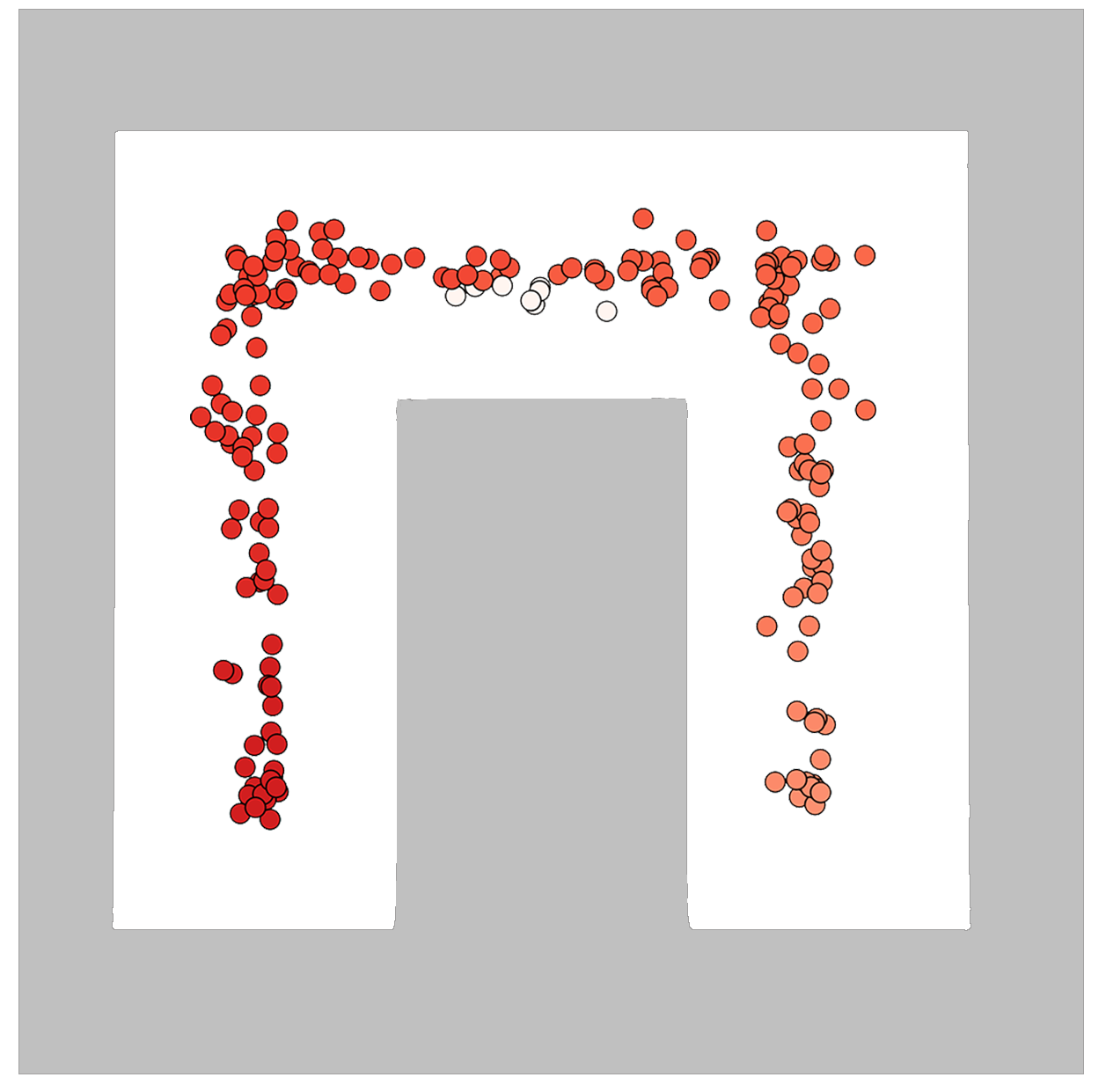}
    \\
    \includegraphics[width=.24\textwidth]{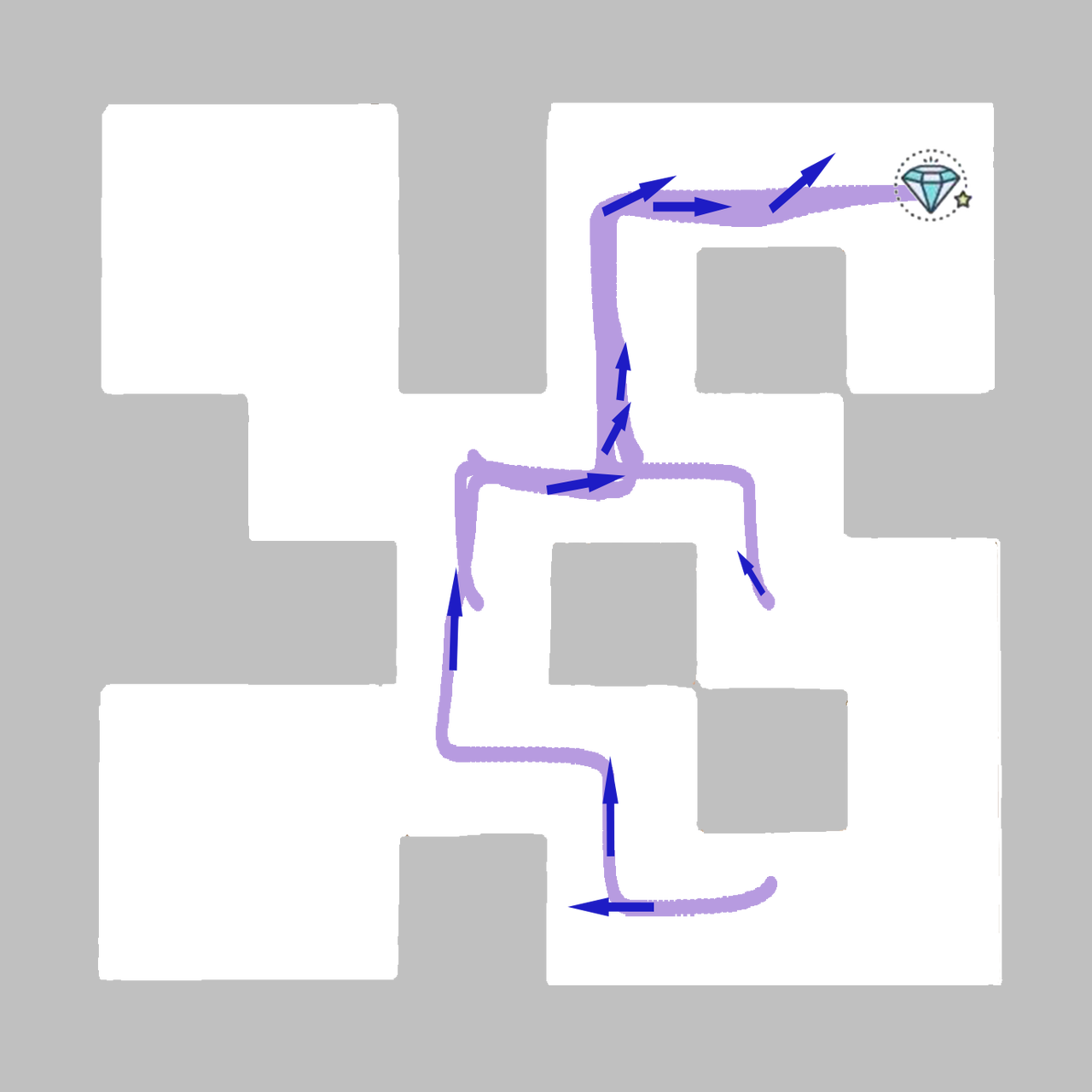}
    \includegraphics[width=.24\textwidth]{figure/forward_medium_return.PNG}
    \includegraphics[width=.24\textwidth]{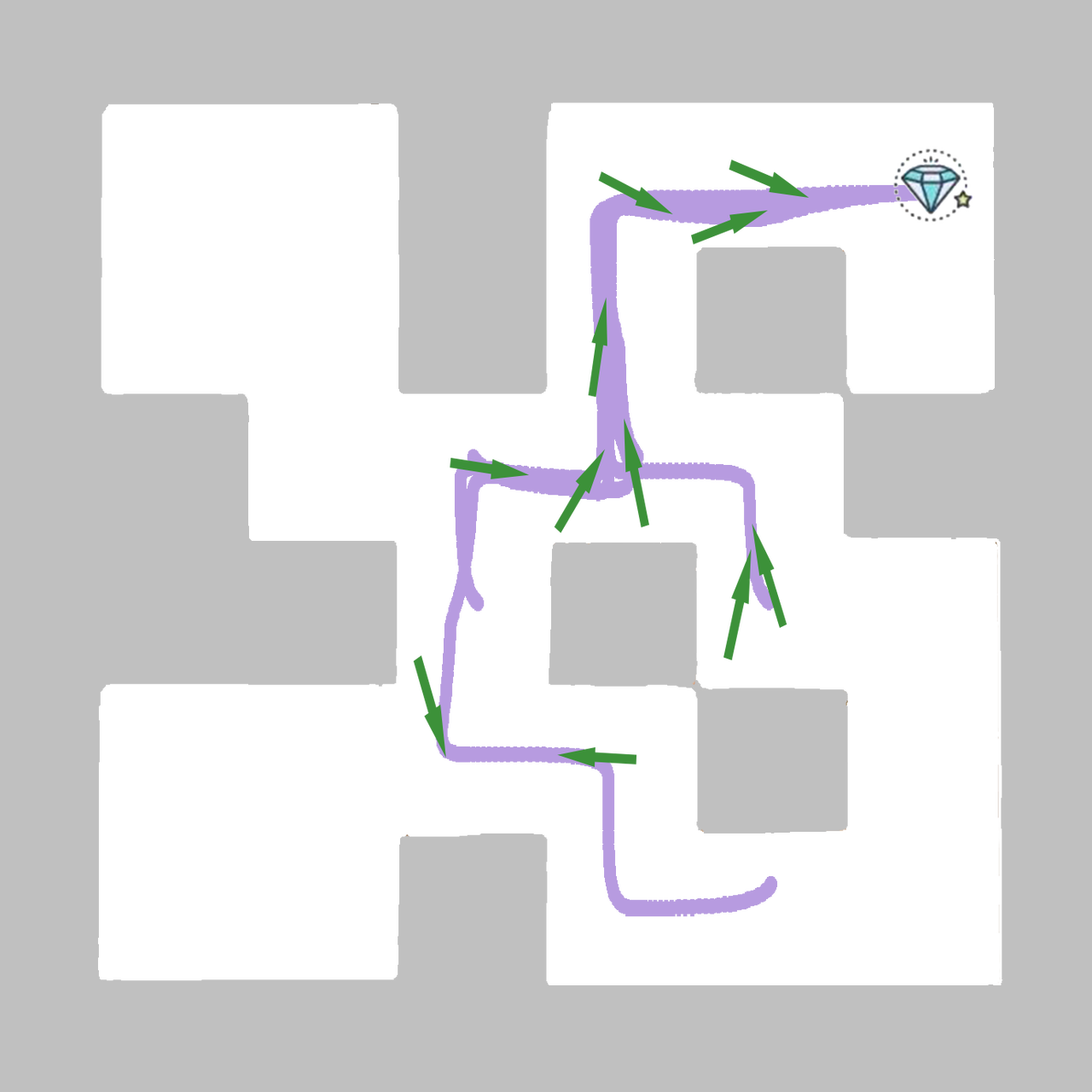}
    \includegraphics[width=.24\textwidth]{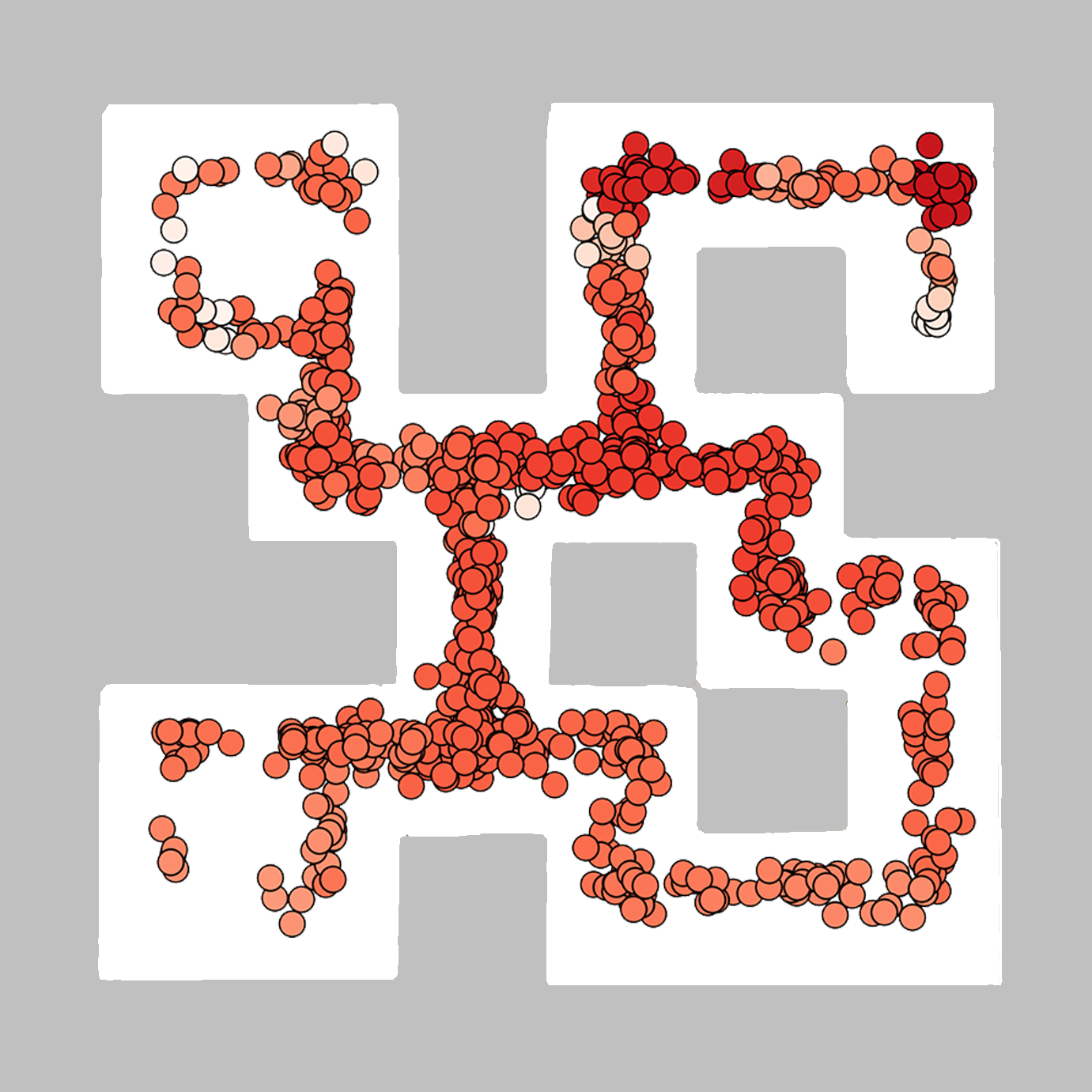}
    \\
    \subfigure[Forward rollout of MILO]{
    \includegraphics[width=.24\textwidth]{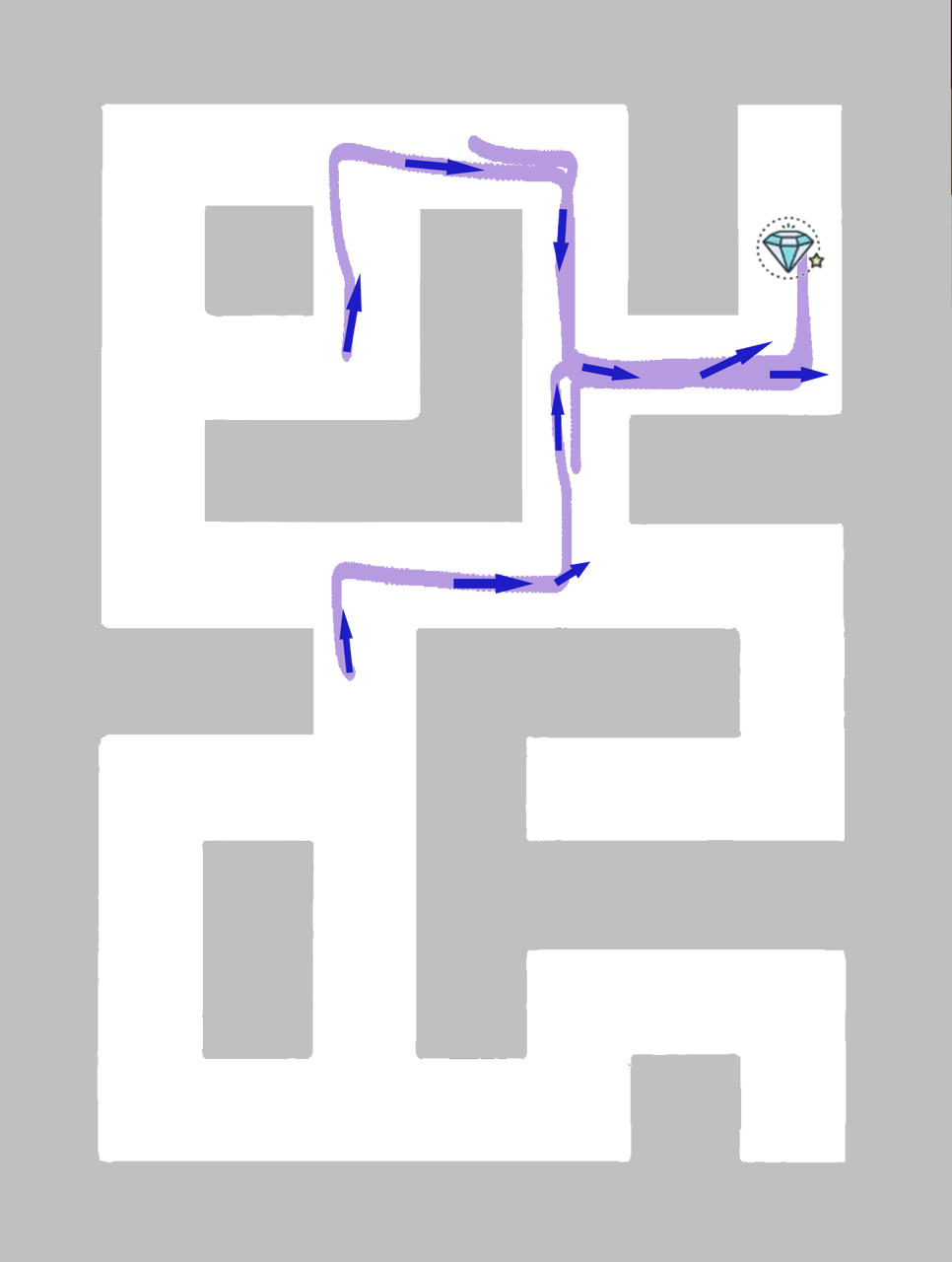}
    \label{forward_data}
    \hspace{-0.2cm}
    }
    \subfigure[Cumulative return of MILO]{
    \includegraphics[width=.24\textwidth]{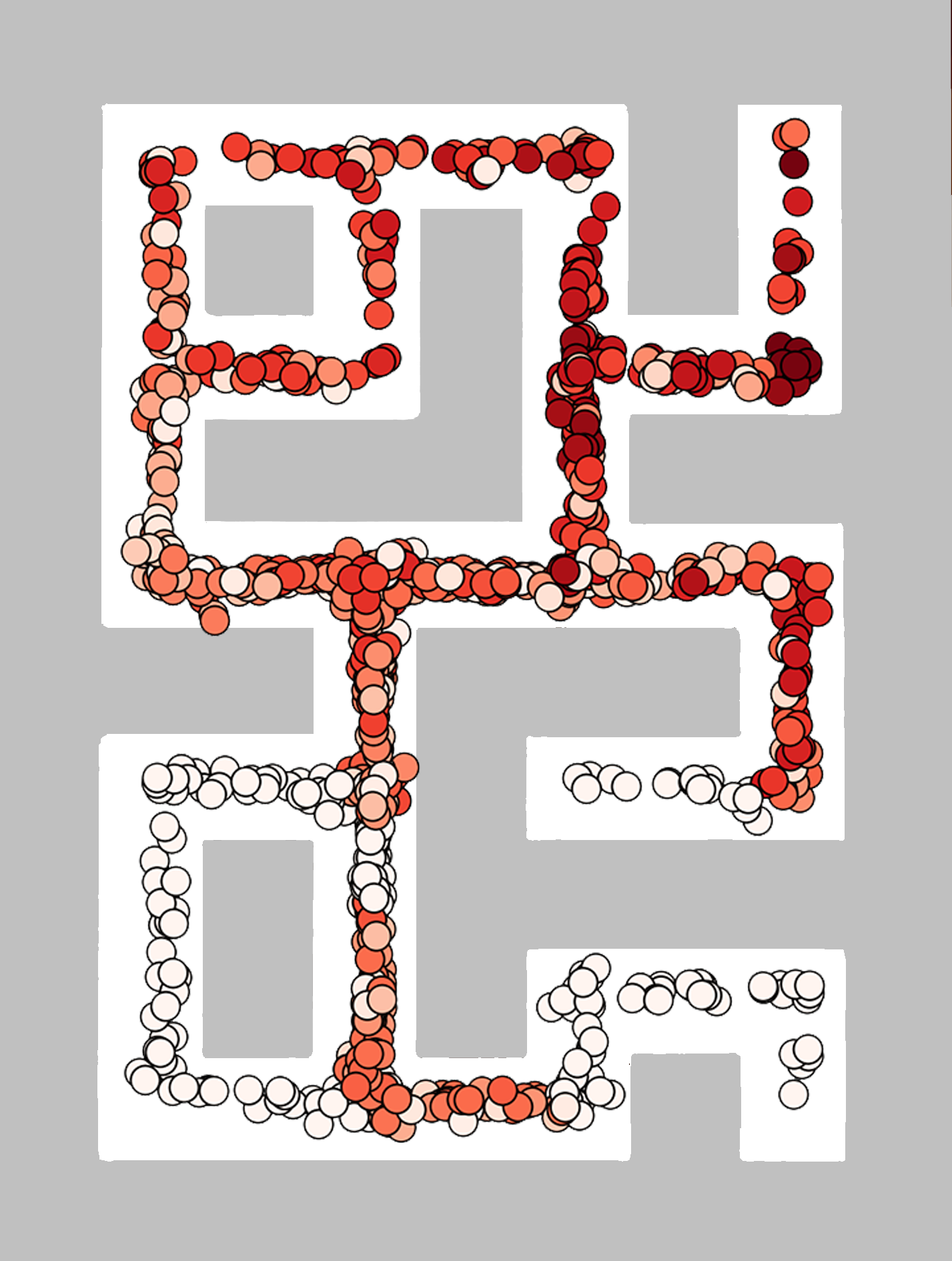}
    \label{forward_performance}
    \hspace{-0.2cm}
    }
    \subfigure[Reverse rollout of SRA]{
    \includegraphics[width=.24\textwidth]{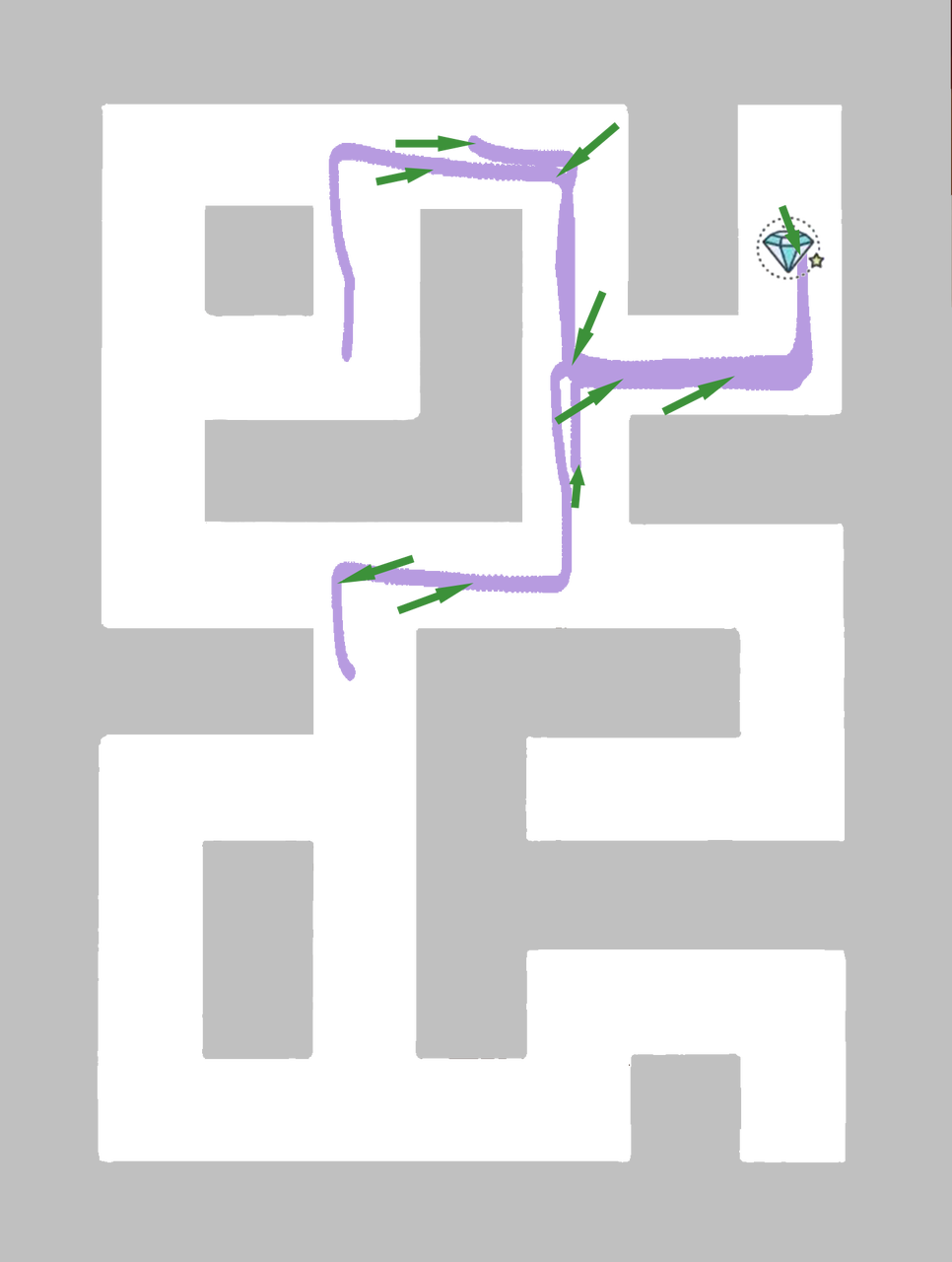}
    \label{reverse_data}
    \hspace{-0.2cm}
    }
    \subfigure[Cumulative return of SRA]{
    \includegraphics[width=.24\textwidth]{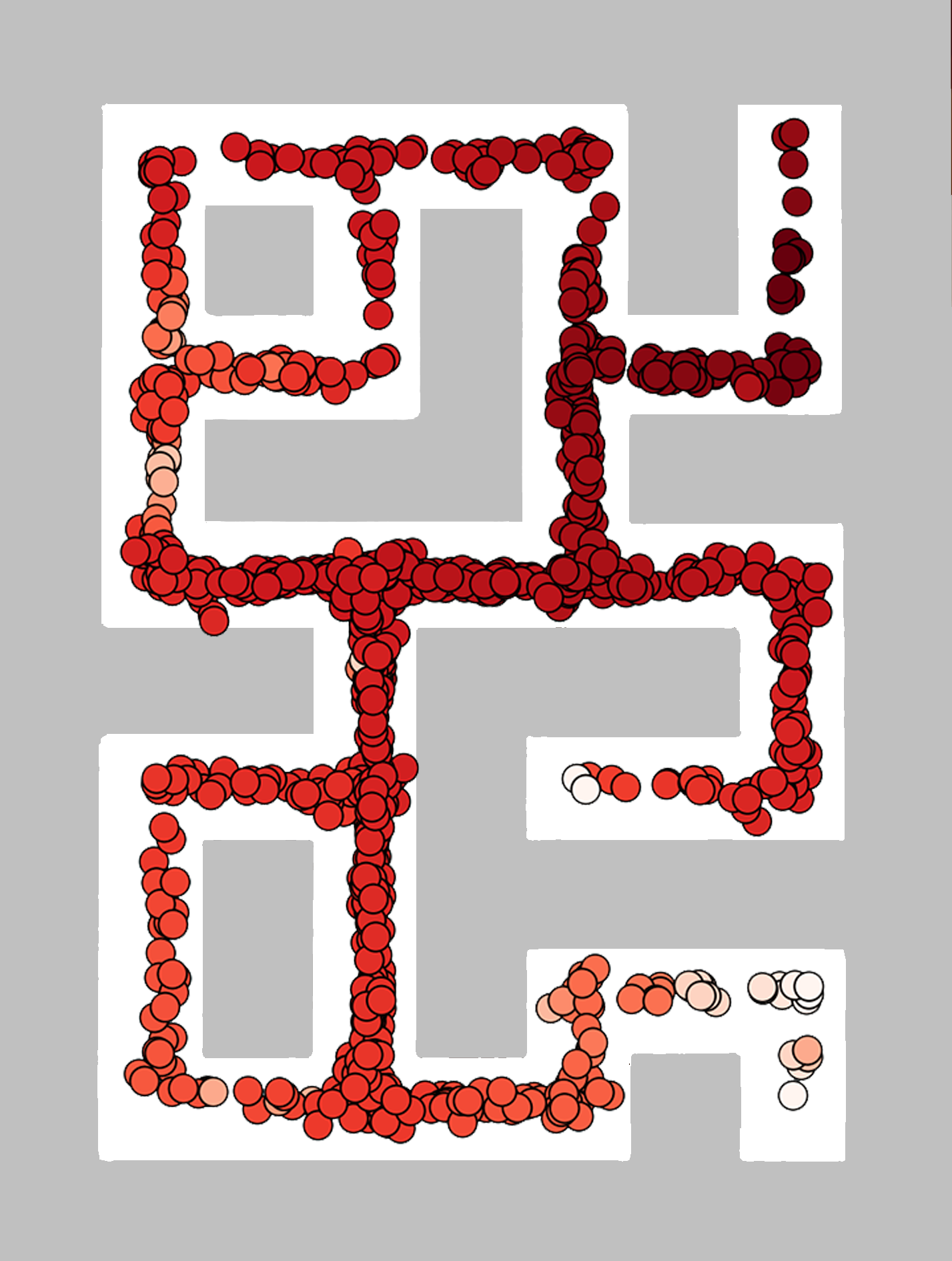}
    \label{reverse_performance}}
    \caption{Forward Augmentation v.s. Reverse Augmentation. The visualization on navigation tasks. }
  \end{figure*} 
  
  From the results, we could find that at the beginning of learning, the ability of the policy is relatively weak, and the data obtained by the reverse model is mainly concentrated near the expert trajectories. However, as the ability of the policy in the states near the expert trajectories is improved, the data obtained by SRA sampling contains more diverse expert-unobserved states. In this process, the improvement of policy generalization ability and SRA's exploration of expert-unobserved states help each other. Eventually, this allows the policy to achieve higher cumulative return in almost all areas of the medium maze. 

  In addition to the visualization analysis, we also provide a quantitative 
  ablation study with the self-paced module in Table~\ref{ablation}. We can find that in tasks like locomotion where the expert distribution is relatively narrow, the self-paced process can also effectively generate data to improve performance better. These results could further support the effectiveness of the proposed self-paced module.

  \begin{table*}[tb]
    \caption{Combination of SRA and different offline reinforcement learning methods.}
    \label{varying_orl}
    \setlength{\tabcolsep}{1.5mm}{
    \begin{tabular}{c|c|c|c|c|c|c|c|c}
      \hline\hline
      DataSet & IQL & SRA+IQL & TD3BC & SRA+TD3BC & AWAC & SRA+AWAC & SAC-N & SRA+SAC-N  \\
      \hline
      maze2d-umaze-sparse-v1 & 64.9$\pm$8.51 & 155.$\pm$6.20 $\uparrow$ & 38.1$\pm$12.9 & 145.$\pm$7.27 $\uparrow$ & 68.6$\pm$14.5 & 135.$\pm$10.1 $\uparrow$ & 151.$\pm$6.44 & 150.$\pm$6.47            \\
      maze2d-medium-sparse-v1 & 83.0$\pm$8.84 & 147.$\pm$5.67 $\uparrow$ & 22.4$\pm$9.64 & 140.$\pm$8.55 $\uparrow$ & 100.$\pm$13.5 & 87.8$\pm$16.4           & 147.$\pm$10.3 & 153.$\pm$7.36 $\uparrow$\\
      maze2d-large-sparse-v1 & 108.$\pm$16.7 & 150.$\pm$14.9 $\uparrow$ & 57.9$\pm$13.2 & 143.$\pm$17.7 $\uparrow$ & 74.8$\pm$16.8 & 89.9$\pm$24.1 $\uparrow$& 128.$\pm$20.7 & 158.$\pm$18.0 $\uparrow$\\
      \hline 
      hopper-medium & 59.5$\pm$4.51 & 90.2$\pm$4.93 $\uparrow$ & 57.2$\pm$1.90 & 96.4$\pm$5.74 $\uparrow$ & 38.3$\pm$3.88 & 85.6$\pm$6.04 $\uparrow$ & 3.33$\pm$0.49 & 107.$\pm$2.31 $\uparrow$\\
      halfcheetah-medium & 43.6$\pm$5.15 & 43.7$\pm$1.72 $\uparrow$ & 43.2$\pm$0.82 & 1.30$\pm$1.25 & 42.0$\pm$1.77 & 44.9$\pm$1.95 $\uparrow$ & -.15$\pm$0.08 & 7.49$\pm$2.71 $\uparrow$\\
      walker2d-medium & 97.6$\pm$2.85 & 101.$\pm$3.60 $\uparrow$ & 89.6$\pm$3.40 & 103. $\pm$3.58 $\uparrow$ & 90.8$\pm$5.91 & 94.3$\pm$4.46 $\uparrow$ & 4.19$\pm$0.42 & 86.5$\pm$7.72 $\uparrow$\\
      ant-medium & 87.3$\pm$5.10 & 88.9$\pm$7.18 $\uparrow$ & 90.4$\pm$5.42 & 42.0$\pm$8.30 & 57.0$\pm$8.39 & 82.2$\pm$9.29 $\uparrow$ & -27.$\pm$3.40 & 47.2$\pm$7.43 $\uparrow$\\
    \hline
    Win/Tie/Loss & \multicolumn{2}{c|}{7/0/0}&\multicolumn{2}{c|}{5/0/2}&\multicolumn{2}{c|}{6/0/1}&\multicolumn{2}{c|}{6/0/1}\\
    \hline\hline
  \end{tabular}}
  \end{table*}

  \subsection{Concerns on Model-based Augmentation}

  The key difference between our work and previous model-based offline imitation learning methods~\cite{MILO,CLARE} is that we have used a reverse model instead of the previous forward model. 
  To delve deeper into understanding this difference, in this subsection, we provide a detailed visual demonstration in the navigation domain (Maze2D-sparse), where the covariate shift problem is crucial. 

  Figure~\ref{forward_data} shows the expert demonstrations and the augmented trajectories by forward models of MILO~\cite{MILO}. 
  We could find a dilemma, the longer the trajectory of the forward model rollout, the further it is from the expert-observed states, thus losing the guidance of policy learning. Existing methods, like MILO and CLARE, constrain their policies close to the expert trajectories to reduce the occurrence of this kind of situation. However, it's hard for these kinds of methods to learn the behaviors out of the expert support. 
  The evaluation of cumulative return in the Figure~\ref{forward_performance} also indicates this limitation. The Figure~\ref{forward_performance} demonstrates the state-wise subsequent cumulative return of MILO. We could find that in the states near the expert trajectories, the subsequent cumulative return is relatively high, which means that when the agent starts from these points, it can obtain higher cumulative returns in the future, that is to say, it can reach the goal. In states far from the expert trajectories, we can see that the cumulative return of the MILO agent tends to be close to 0, showing its limitations. 

  In contrast, our reverse model-based method encourage to explore the expert-unobserved states, as shown in the Figure~\ref{reverse_data}. 
  We could find that the generated trajectories extend from expert-unobserved states to expert-observed states, while providing behavioral samples that improve long-term return for these expert-unobserved states. The longer the trajectory of the reverse model rollout, the more expert-unobserved states can find the path to expert-observed states, thereby improving their long-term return and expanding the feasible state coverage of the learned agents. 

  The state-wise cumulative return of our method is also demonstrated in the Figure~\ref{reverse_performance}. 
  It could be found that our SRA provides a significant improvement on the expert-unobserved states, which support our claim to explore diverse expert-unobserved states. 
  Even in the large maze, our policy can start from the vast majority of states to reach the goal, far beyond the support region provided by the expert demonstrations.

  \subsection{Scalability for Different RL Methods}
  As we discussed in the Subsection~\ref{subsection34}, our SRA is agnostic to the specific offline reinforcement learning methods. In this subsection, we validate the scalability via pairing with the different offline reinforcement learning methods, includes IQL~\cite{IQL}, TD3BC~\cite{TD3BC}, AWAC~\cite{AWAC} and SAC-N~\cite{SACN}. 
  We wrap the original offline reinforcement learning methods into the offline imitation learning setting through UDS, that is, we label the expert data and supplementary data as 1 and 0 respectively. Under the same hyper-parameters, we add SRA data augmentation for method comparison. 
  The results on both navigation and locomotion tasks are reported in Table~\ref{varying_orl}. 
  We could find that our data augmentation module, SRA, consistently improves these baseline methods and show stronger robustness. 
  Specifically, as an advanced offline reinforcement learning method, IQL can effectively alleviate the overestimation of Q-learning. Its performance is relatively stable in both SRA-equipped and non-SRA-equipped situations. Therefore, it is also what we suggest as a basic method in the SRA framework. 
  Overall, the results across four popular offline reinforcement learning methods clearly support the scalability of SRA, demonstrating its superiority. 

\section{Conclusion}

In this paper, we study the covariate shift problem of offline imitation learning. 
The key difficulty is that it is challenging for the agent to obtain trustworthy behavior on expert-unobserved states for policy optimization. To overcome this issue, we present a novel framework, offline imitation learning with Self-paced Reverse Augmentation. 
This framework generates the trajectories from expert-unobserved states to expert-observed states in a self-paced way. When the agent encounters these expert-unobserved states, it can follow the generated trajectory to reach the expert-observed states, thereby improving the long-term return. 
To the best of our knowledge, this is the first time to introduce the reverse data augmentation to the offline imitation learning. It is different from previous methods based on the forward model. That is, it allows the strategy to explore more diverse expert-unobserved states. 
In the empirical studies, the effectiveness of our Self-paced Reverse Augmentation has been verified in a series of benchmark tasks. 
Not only has it achieved state-of-the-art performance, but it has also offered behavioral guidance and enhanced capabilities in the expert-unobserved states, providing a promising way to mitigate the covariate shift of offline imitation learning. 

This work is inspired by the BCDP~\cite{BCDP}, which presents the idea of leading the agent from expert-unobserved states to expert-observed states. We propose a reverse-model-based solution to generate diverse trajectories from expert-unobserved states to the expert-observed states. This strategy has a significant advantage in mitigating covariate shifts compared to previous forward-model-based methods. 
A concurrent work, ILID~\cite{ILID}, presents a similar idea and proposes a model-free data selection method leading the agents to focus on the trajectories whose resultant states fall within the expert data manifold. 
Further exploration of model-free strategies and the unified solutions with our model-based framework is an interesting direction. 
Another potential future direction is to extend Self-paced Reverse Augmentation with advanced model learning methods, such as dynamic quantization, to further improve the quality of augmented trajectories. 

\begin{acks}
  This research was supported by Leading-edge Technology Program of Jiangsu Science Foundation (BK20232003), National Science Foundation of China (62176118) and the Postgraduate Research \& Practice Innovation Program of Jiangsu Province (KYCX24\_0233).
\end{acks}

\bibliographystyle{ACM-Reference-Format}
\balance
\bibliography{kdd24}


\end{document}